\pdfoutput=1
\documentclass[11pt]{article}

\usepackage[final]{acl}

\usepackage{times}
\usepackage{latexsym}

\usepackage[T1]{fontenc}
\usepackage[utf8]{inputenc}
\usepackage{microtype}
\usepackage{inconsolata}
\usepackage{graphicx}
\graphicspath{{fig/}}
\usepackage[most]{tcolorbox}
\usepackage{listings}
\usepackage{booktabs}
\usepackage{multirow}
\usepackage{xcolor}

\definecolor{codebg}{HTML}{F7F7F7}
\definecolor{codekey}{HTML}{0000AA}
\definecolor{codestr}{HTML}{008800}
\definecolor{codecom}{HTML}{888888}
\definecolor{codefn}{HTML}{AA5500}

\lstdefinestyle{pytool}{
  language=Python,
  basicstyle=\ttfamily\scriptsize,
  keywordstyle=\color{codekey}\bfseries,
  stringstyle=\color{codestr},
  commentstyle=\color{codecom}\itshape,
  emphstyle=\color{codefn}\bfseries,
  emph={execute,_authority_score,_pick_candidates_from_lens,
        _slug_to_condition,_is_valid_condition,_post},
  backgroundcolor=\color{codebg},
  breaklines=true,
  breakatwhitespace=false,
  columns=fullflexible,
  keepspaces=true,
  showstringspaces=false,
  frame=none,
  xleftmargin=2pt,
  xrightmargin=2pt,
  aboveskip=2pt,
  belowskip=2pt,
  tabsize=4,
}
\usepackage{epigraph}
\usepackage{enumitem}
\usepackage{xspace}

\newcommand{\methodname}{\textsc{ToolSmith}\xspace}

\newcommand{\dup}[1]{\textcolor{green!60!black}{$+$#1}}
\newcommand{\ddn}[1]{\textcolor{red!80!black}{$-$#1}}

\definecolor{toolhlbg}{HTML}{D6E4F5}
\definecolor{toolhlfg}{HTML}{1F3A7A}
\newtcbox{\toolhl}{on line, arc=3pt, colback=toolhlbg, colframe=toolhlbg,
  boxrule=0pt, boxsep=0pt, left=4pt, right=4pt, top=1.5pt, bottom=1.5pt,
  fontupper=\bfseries\color{toolhlfg}}

\title{\raisebox{-0.12ex}{\includegraphics[height=1.65ex]{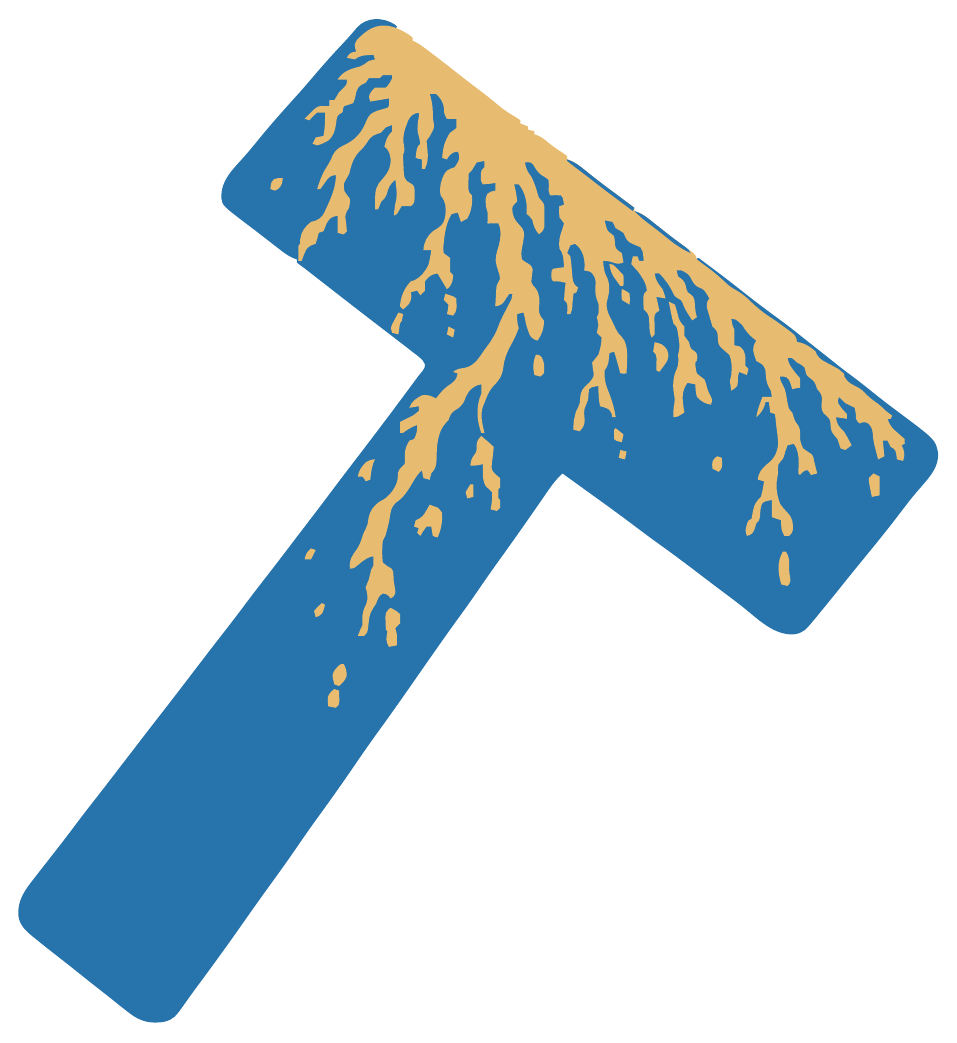}}\hspace{0.25em}\methodname\\
An Automatic Tool-Generation Framework for Multimodal Search Agents}
\author{Rui Sun\textsuperscript{$\spadesuit$}, Junhua Huang\textsuperscript{$\spadesuit$}, Luca Zancato\textsuperscript{$\clubsuit$}, Md Mofijul Islam\textsuperscript{$\clubsuit\dagger$}, Kai-Wei Chang\textsuperscript{$\spadesuit\dagger$}
\\
\textsuperscript{$\spadesuit$} University of California, Los Angeles \quad
\textsuperscript{$\clubsuit$} AWS GenAI \quad \\
\texttt{ruis@g.ucla.edu}
}

\newcommand{\blfootnote}[1]{%
  \begingroup
  \renewcommand\thefootnote{}\footnote{#1}%
  \addtocounter{footnote}{-1}%
  \endgroup
}

\begin{document}
\maketitle
\blfootnote{\textsuperscript{$\dagger$} Equal advising.}

\begin{abstract}
Recent multimodal search agents use reasoning, vision, and live web search across many turns to answer questions beyond parametric knowledge by orchestrating a small library of hand-crafted search and visual tools. However, the library is laborious to create and maintain, and frozen when new query types, evidence formats, and failure modes are encountered in the wild. Human practitioners, by contrast, cope with insufficient tools by fashioning new ones from familiar parts and reusing them across tasks. Inspired by this, we propose \methodname, an automatic tool-generation framework that grows and reuses modules on top of a lean base tool library. We evaluate \methodname on three in-domain multimodal search benchmarks and ten out-of-domain benchmarks under a VLM-as-a-Judge protocol. \methodname consistently improves over the base library across deployed agents of varying scales, with larger backbones reaping greater gains, demonstrating the scalability of the approach. Our code and data is available at \url{https://threesr.github.io/toolsmith/}.
\end{abstract}

\begin{figure}[t!]
\centering
\includegraphics[width=1.0\columnwidth]{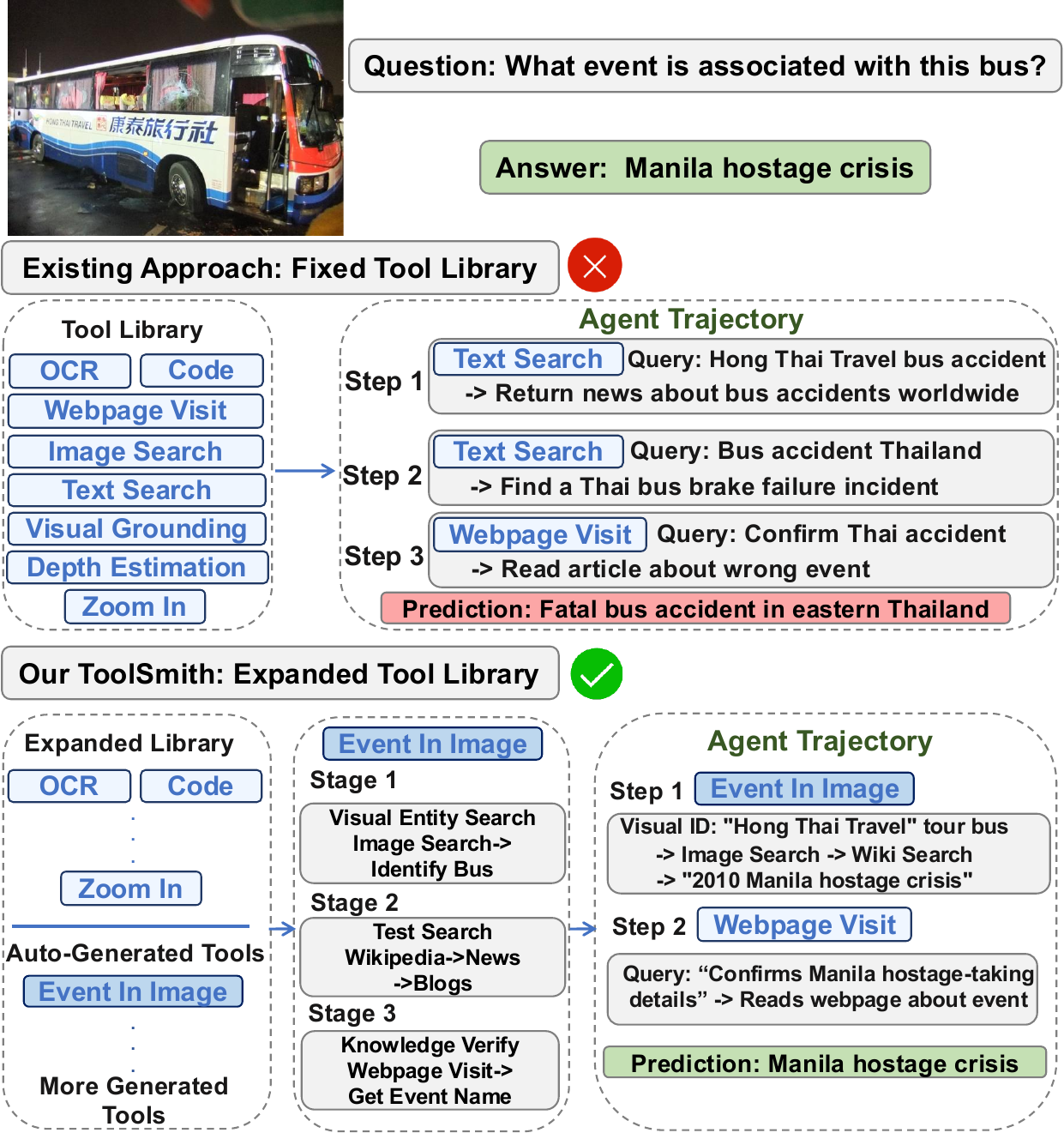}
\caption{We propose an automatic tool-generation framework for multimodal search agents. In the example shown, an agent restricted to a \emph{fixed} tool library is bound to its pre-defined primitives and fails on queries that no original tool can directly resolve. Our framework instead \emph{grows} the library: the agent operates over an extended tool library that includes newly generated tools and can invoke them freely when solving real queries. Here, the newly generated \toolhl{Event In Image} tool resolves the query through finer-grained information verification and image search.}
\label{fig:teaser}
\end{figure}

\section{Introduction}
\label{sec:intro}

Multimodal search agents \citep{geng2026webwatcher,huang2026vision,yao2026mm,chen2026opensearch,wu2025mmsearch,narayan2025deepmmsearch} answer information-seeking questions beyond static knowledge inside the model, such as identifying the source of the rare knowledge, cross-checking a numerical claim against the live web, or assembling a multi-step provenance argument across images and retrieved pages. Contemporary systems handle such tasks by orchestrating reasoning, perception, and live web search across many turns over a fixed inventory of hand-crafted tools. It typically contains web search, webpage fetching, image grounding, OCR, depth estimation, and code execution. Throughout this paper, we use \emph{tool} to mean a callable Python utility, exposed to the agent through a typed schema, that takes structured input (a text query, an image, a URL) and returns evidence the agent's reasoning loop consumes in a subsequent turn; \emph{generating a new tool} means producing a new such callable schema, execution logic, and all, and admitting it to the catalog the agent draws from at inference.

Nevertheless, hand-crafted tools suffer from inherent limitations. First, tool inventories shipped with production systems and developer-curated skill banks \citep{openclawskills} are expensive to maintain and frozen at the developer's anticipated coverage. When new query types, evidence formats, and failure modes are encountered in the wild, human intervention will be needed to modify the tools. Second, complementary lines of work that learn memory \emph{over} a fixed inventory \citep{xiao2025toolmem} or train an agent to use existing tools more efficiently \citep{gao2025multi,wu2026vtool} sharpen \emph{how} the agent uses its tools but do not expand \emph{which} tools it has.

A natural alternative is letting the agent itself \emph{generate} the missing tool. Tool generation has been studied across adjacent domains such as vision--language understanding, math and table QA, general NLP, and embodied / computer-use agents \citep{chen2024genome,yuan2024craft,wolflein2025llm,wang2023voyager,wang2025inducing}, but has not been carried over to multimodal search, where the inventory remains hand-authored. The transfer is non-trivial: search-agent trajectories are long, and agent state mixes images, link references, and tool outputs across turns.

When exploring the world, human practitioners both use existing tools and fashion new ones when existing tools fall short \citep{goodall1964,beck1980,tomasello1999}. We argue that a multimodal search agent should be equipped with the same capability. When existing tools fall short, the agent should organically \emph{fashion new ones} from familiar parts, refining and reusing them across tasks. Currently, multimodal search agents only share the \emph{using} half of this picture but conspicuously not the \emph{fashioning} half. When their tool set fails, they 'let it go' rather than enlarge and revise the toolbox. Moreover, humans do not blindly adopt every novel implementation they sketch. They \emph{verify} it on test cases before adding it to the working library \citep{schon1983}. An agent equipped to grow its tools must also be equipped to verify them.

To this end, we introduce \methodname, an automatic tool-generation framework organized as four model-agnostic phases: 1) \emph{Trajectory Mining}, 2) \emph{Reuse-or-Grow Decision}, 3) \emph{Tool Generation}, 4) \emph{Two-Gate Verification}. We cluster the agent's failed trajectories into capability gaps \citep{shinn2023reflexion,andrychowicz2017hindsight,chen2025scaling}, check each gap against a lean library of eight primitives to decide reuse versus generation, prompt a generator VLM to author new tools as Python source under structural and anti-hardcode constraints, and admit candidates only after sandboxed sanity and LLM static code-review gates.

We evaluate \methodname across 3 in-domain multimodal search benchmarks (FVQA, InfoSeek, EncyclopedicVQA) and 10 out-of-domain benchmarks under a VLM-as-a-Judge manner. When using the Qwen3-VL-8B-Instruct model, admitted tools improve the base tool library by \textbf{+2.20pp} in-domain and \textbf{+1.33pp} out-of-domain tasks. Gains scale with backbone capacity, reaching \textbf{+2.98pp} out-of-domain on Qwen3-VL-32B-Instruct and a consistent uplift on GPT-5. We also utilize a comprehensive ablation study to demonstrate how different tool generation models and training samples affect the benchmark performance. Our code and data will be publicly available.

\paragraph{Contributions.}
\begin{enumerate}[leftmargin=*,noitemsep,topsep=2pt,parsep=0pt]
    \item We propose \methodname, to the best of our knowledge, the first \emph{automatic tool-generation} framework for multimodal search agents, with a model-agnostic four-phase pipeline over a base tool library of search, visual, and code-execution tools.
    \item On 3 in-domain and 10 out-of-domain multimodal search benchmarks, admitted tools improve the base library across different models, with gains extending to larger backbones, which demonstrate the scalability and flexibility of our proposed \methodname.
    \item We conduct ablation studies to analyze overhead, new-tool usage, tool-call distribution, tool-filtering, and failure modes, which provides an objective landscape of \methodname.
\end{enumerate}

\section{Related Work}
\label{sec:related}

\paragraph{Multimodal search agents.} Our work targets multimodal search, where a query is resolved by orchestrating reasoning, perception, and live web retrieval across many turns. A growing collection of benchmarks frames this task: MMSearch \citep{jiang2025mmsearch} couples a multimodal query with a curated search engine; MMSearch-Plus \citep{tao2026mmsearch} adds provenance-aware verification; BrowseComp-VL \citep{geng2026webwatcher} and MM-BrowseComp \citep{li2025mm} raise cross-modal information-seeking difficulty; VisBrowse-Bench \citep{zhang2026visbrowse} targets visual-native search; and RealX-Bench \citep{hong2026deepeyesv2} couples fine-grained perception with multi-hop retrieval. Beyond pure search, the landscape extends to embodied tasks that integrate web information with physical action \citep{hong2026embodied}. These tasks build on vision-language understanding, where models must ground fine-grained visual evidence \citep{you-etal-2022-find,sun-etal-2023-unifine,wang2024journeybench} and abstain when the visual context is insufficient \citep{liu2024detecting}. Recent browsing agents tackle these benchmarks via hand-engineered tool inventories \citep{geng2026webwatcher,huang2026vision,yao2026mm,chen2026opensearch,wu2025mmsearch,narayan2025deepmmsearch,zhou2026openvlharness}, supervised fine-tuning or reinforcement learning that teaches tool use and action selection \citep{jiang2026agent}, or memory learned over a fixed inventory \citep{xiao2025toolmem,hasegawa2025tama,jiang2026memory}. Our contribution is orthogonal: rather than freeze the inventory, we \emph{generate} it.

\paragraph{Tool generation.} Building on a lineage of LLM-orchestrated modular vision-language reasoning \citep{you2023idealgpt}, our work is closely related to a recent line of methods in which the agent itself constructs new tools. GENOME \citep{chen2024genome} grows a library of vision-module wrappers proposed by an LLM and validated on few-shot examples. CRAFT \citep{yuan2024craft} creates and retrieves Python tools for math, table QA, and image classification. TOOLMAKER \citep{wolflein2025llm} turns code repositories into LLM-callable utilities for general NLP. Voyager \citep{wang2023voyager} grows an executable skill library for an open-ended Minecraft agent, with a parallel line of work \citep{wang2025inducing} carrying the idea to web agents via online programmatic skill induction. EvolveR \citep{wuinteractions} distills past trajectories into textual principles that guide future behaviour, and WorldMind \citep{ren2026aligning} builds a symbolic world-knowledge repository from agent experience for physically-grounded action. Compared with these, we introduce the tool generation framework into the multimodal search domain.

\section{Method}
\label{sec:method}

\begin{figure*}[t]
\centering
\includegraphics[width=\textwidth]{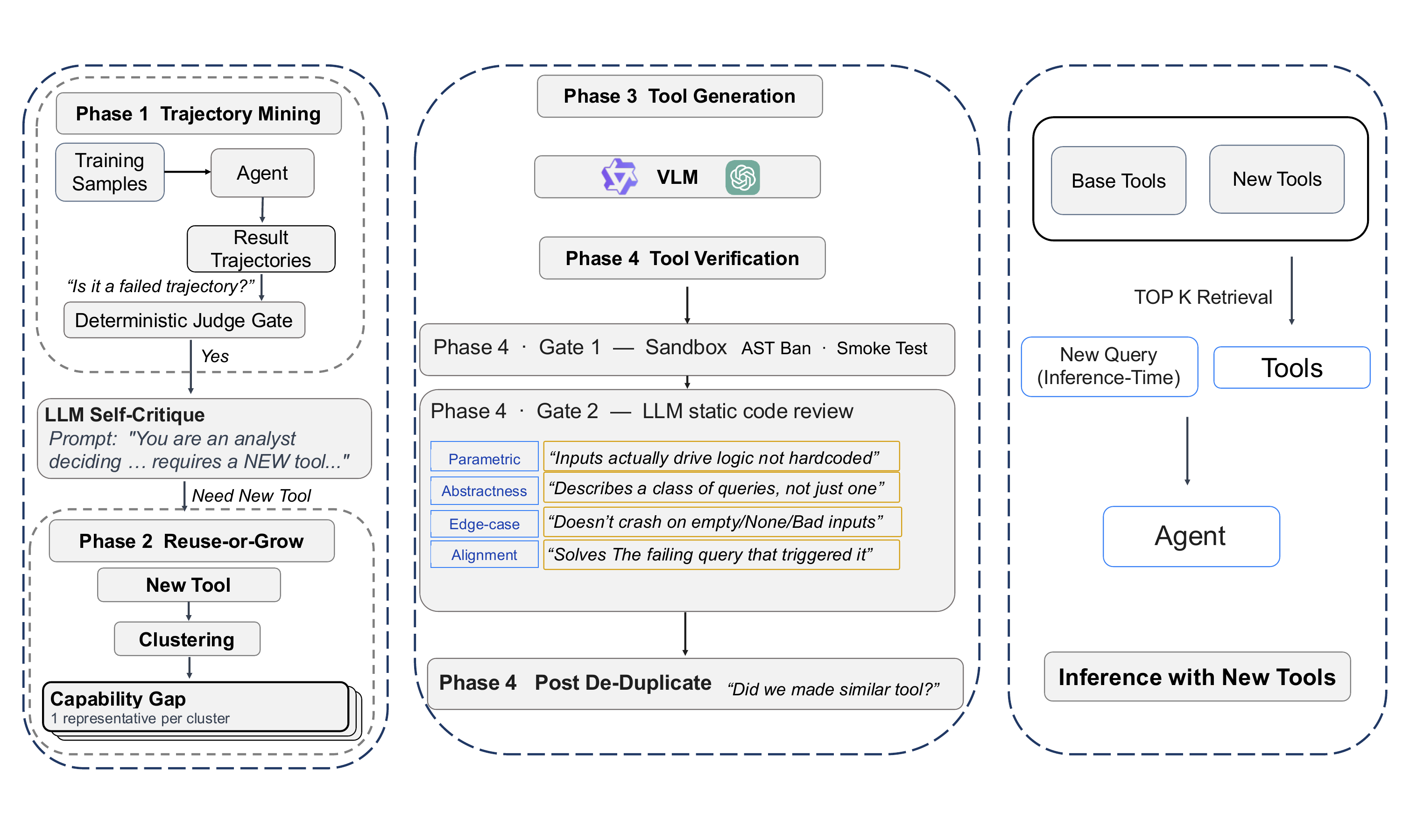}
\caption{Overview of the \methodname pipeline: capability gaps mined from agent trajectories on a training pool are filtered through reuse-or-grow against the tool catalog, passed to an LLM that authors each new tool as Python source, and admitted only after verification gates with a feedback-driven retry on rejection.}
\label{fig:framework}
\end{figure*}

\methodname is organized as a four-phase pipeline (Figure~\ref{fig:framework}): Phase~1 mines \emph{capability gaps} (query patterns the base tools cannot handle) from the baseline agent's failed trajectories. Phase~2 decides, per gap, whether to reuse an existing tool or escalate to generation. Phase~3 prompts a generator LLM to author the new tool as Python code. Phase~4 admits the candidate after verification gates. At inference time, a lightweight retrieval step selects relevant generated tools per query.

\paragraph{Notation.} Let $\mathcal{A}_\theta$ be the baseline agent, $\mathcal{D}_{\text{train}}$ the training pool, and $\mathcal{T}_t$ the tool catalog at round~$t$, initialized with the eight base tools, $\mathcal{T}_0=\mathcal{T}_{\text{base}}$. A trajectory $\tau$ records the agent's full tool-call sequence on a single query. We write $G_\phi$ for the generator LLM, $\psi$ for a sentence encoder, and $\cos(\cdot,\cdot)$ for cosine similarity.

\subsection{Phase 1: Trajectory Mining}
\label{sec:phase1}

We roll out $\mathcal{A}_\theta$ on $\mathcal{D}_{\text{train}}$ and apply a two-layer trigger to each trajectory. \textbf{Layer~A} (cheap signal) flags any trajectory whose final answer the LLM judge marks as incorrect. \textbf{Layer~B} (LLM self-critique) then reads each flagged trajectory and assigns a failure label from \{\textsc{new\_tool\_needed}, \textsc{argument\_mistake}, \textsc{search\_noise}, \textsc{ambiguous}\}. Only \textsc{new\_tool\_needed} trajectories yield a capability gap, which we record as the tuple of (failing query, tools the agent already tried, the self-critique's failure reason, and an image reference when image-grounded).

\paragraph{Clustering.} To ensure diversity in candidates, we embed each capability gap as the concatenation of its question, prediction, and gold answer using a sentence encoder~$\psi$, then apply agglomerative clustering (average linkage) over cosine distance into $C$ clusters. One representative per cluster enters Phases~2--4.

\subsection{Phase 2: Reuse-or-Grow Decision}
\label{sec:phase2}

Phase~2 decides, for each capability gap, whether the current tool catalog already covers it or a new tool is needed. We compute the TF--IDF cosine similarity between the gap's text (its query and subtask) and every existing tool's description, and take the maximum:
\begin{equation}
  \rho_{\text{pre}}(g, \mathcal{T}_t) = \max_{t\in\mathcal{T}_t}\cos\!\bigl(\mathrm{tfidf}(\widetilde{g}),\,\mathrm{tfidf}(d_t)\bigr),
  \label{eq:reuse}
\end{equation}
where $d_t$ is the description of tool $t$ and $\widetilde{g}=q_g\oplus s_g$ concatenates the gap's query and subtask. If $\rho_{\text{pre}}>\tau_{\mathrm{pre}}$, the gap is treated as covered by the best-matching tool and skipped; otherwise it is forwarded to Phase~3. TF--IDF suffices here because tool descriptions are short, the vocabulary is domain-specific (search, image, OCR, code), and we only need to detect obvious overlaps.

\subsection{Phase 3: Tool Generation}
\label{sec:phase3}

Phase~3 takes each gap that Phase~2 forwards and writes a new tool for it. We prompt a generator LLM $G_\phi$ to produce Python source code conditioned on the gap, the current catalog, a system prompt that specifies the tool-class interface and forbids dynamic execution / unsanctioned I/O / hard-coded answers, and a handful of example tools the generator can imitate (full prompts and examples in Appendix~\ref{sec:toolgen_prompts}). The output is a candidate $c$ comprising Python source, a typed schema, and a short natural-language description, which is then forwarded to Phase~4 for verification. Representative code listings of admitted, failed, and filtered candidates appear in Appendix~\ref{sec:tool_examples}.

\subsection{Phase 4: Verification and Inference}
\label{sec:phase4}

A candidate is admitted only if it passes both verification gates followed by a redundancy check against the existing catalog. We run the gates in order of increasing computational cost: Gate~1 is a cheap static check, so candidates that fail it are dropped before we pay for Gate~2's LLM critic.

\paragraph{Gate 1: Sandboxed sanity.} The candidate's source is parsed and scanned for dynamic-execution and dangerous-I/O patterns, then executed once on a minimal dummy input under a ten-second wall-clock budget. Anything that violates the static checks or errors at runtime is rejected.

\paragraph{Gate 2: LLM static code review.} An LLM critic scores the candidate on four 0--5 axes (parametric usage, description abstractness, edge-case handling, and task alignment) and admits it only when the mean score is at least $\mu_{\min}$ and the minimum across axes is at least $a_{\min}$. The critic may use a different model from the generator; tools with overly narrow descriptions or missing error handling are caught here.

\paragraph{Feedback-driven retry.} When either gate rejects a candidate, we re-prompt $G_\phi$ with a short natural-language summary of which gate failed and why, up to a per-candidate budget $A$. Empirically, most retries either pass on the second attempt or remain wrong for the same reason, so $A$ is capped at a small constant.

\paragraph{Tool catalog update.} A candidate that survives both gates is then compared against every existing tool by TF--IDF cosine similarity over descriptions (the same construction as Eq.~\ref{eq:reuse}, with a stricter threshold $\tau_{\mathrm{post}}$). Only candidates whose closest match falls below $\tau_{\mathrm{post}}$ are admitted to the next-round catalog $\mathcal{T}_{t+1}$; redundant duplicates are dropped.

\paragraph{Inference.} Exposing the full generated catalog alongside the base primitives would inflate the per-turn schema and degrade tool selection. We instead retrieve a per-query subset: each tool description~$d_t$ is pre-embedded with~$\psi$; given a test query~$q$, the agent sees
\begin{equation}
  \mathcal{T}_q = \mathcal{T}_{\text{base}} \;\cup\; \operatorname{top\text{-}K}\!\bigl(\cos(\psi(q),\, \psi(d_t))\;;\; t \in \mathcal{T}_{\text{new}}\bigr)
  \label{eq:routing}
\end{equation}
yielding $|\mathcal{T}_{\text{base}}| + K$ tools per sample. The base primitives are always visible (they are referenced by name in the agent's system prompt); the generated tools rotate per query based on relevance. Hyperparameter values are listed in Appendix~\ref{sec:setup_hp}.

\section{Experimental Setup}
\label{sec:setup}

\subsection{Base Tool Library}
\label{sec:setup_base}

Our base library consists of eight primitive tools grouped into three capability categories: \emph{Search}, \emph{Vision}, and \emph{Code Execution}. The search tools are powered by the Serper API\footnote{\url{https://serper.dev/}}; the visual tools wrap SAM3 \citep{carion2026sam}, PaddleOCR-VL \citep{cui2025paddleocr}, and Depth Anything 3 \citep{lin2025depth} as their underlying models. Table~\ref{tab:base_tools} lists the eight tools with a one-line description each. More details can be found at Appendix~\ref{sec:appendix_base_tools}.

\begin{table}[h]
\centering
\small
\begin{tabular}{@{}p{2.6cm}p{4.6cm}@{}}
\toprule
\textbf{Tool} & \textbf{Description} \\
\midrule
\multicolumn{2}{l}{\textit{Search}} \\
Text Search & Web search with automatic query-intent routing. \\
Image Search & Reverse image search (image $\rightarrow$ entities and URLs). \\
Webpage Visit & Fetch URL(s) and extract goal-relevant content. \\
\midrule
\multicolumn{2}{l}{\textit{Vision}} \\
Zoom-in & Crop a region by normalized bounding box. \\
Visual Grounding & Locate regions matching a text query. \\
OCR & Extract printed text from an image. \\
Depth Estimation & Estimate a per-pixel depth map for an image. \\
\midrule
\multicolumn{2}{l}{\textit{Code execution}} \\
Code Execution & Execute a Python snippet (agent-authored or LLM-authored from a goal in plain language) over prior tool outputs. \\
\bottomrule
\end{tabular}
\caption{Base library of eight primitive tools, grouped by capability. Detailed schemas and tool documentation appear in Appendix~\ref{sec:appendix_base_tools}.}
\label{tab:base_tools}
\end{table}

These eight span the major primitive categories a multimodal search agent commonly relies on, providing a broad foundation on top of which our pipeline grows additional, task-specific tools.

\subsection{Tool-Generation Training Pool}
\label{sec:setup_train}

We collect trajectories from the \emph{train} or \emph{val} splits of three knowledge-intensive multimodal search datasets: FVQA \citep{wu2025mmsearch}, InfoSeek \citep{chen2023can}, and EncyclopedicVQA \citep{mensink2023encyclopedic}. The held-out \emph{test} splits of the same three datasets serve as our in-domain (ID) evaluation in Section~\ref{sec:setup_eval}, providing a within-distribution measure of whether the generated tools address the failure modes they were designed for. No example from any out-of-domain benchmark enters tool generation. We defer per-dataset sampling sizes and split choices to Appendix~\ref{sec:appendix_trainsample}.

\subsection{Evaluation Benchmarks}
\label{sec:setup_eval}

We evaluate \methodname along two axes: \emph{in-domain (ID)} performance on the held-out test splits of the three datasets we used to mine capability gaps during training, and \emph{out-of-domain (OOD)} transfer to ten public multimodal search benchmarks unseen during tool generation. Per-benchmark sample counts are summarized in Table~\ref{tab:evalbenches}, and full per-benchmark descriptions are deferred to Appendix~\ref{sec:appendix_eval_benchmarks}.

\paragraph{In-domain (ID).} ID evaluation uses the held-out test splits of FVQA \citep{wu2025mmsearch}, InfoSeek \citep{chen2023can}, and EncyclopedicVQA \citep{mensink2023encyclopedic}. Because the training pool from which we mine capability gaps comes from the \emph{training} splits of these same datasets, the test splits provide a within-distribution measure of how well the generated tools address the failure modes they were designed for. We give full per-benchmark descriptions in Appendix~\ref{sec:appendix_eval_id}.

\paragraph{Out-of-domain (OOD).} OOD evaluation uses ten public multimodal information-seeking benchmarks: MM-BrowseComp \citep{li2025mm}, SimpleVQA \citep{cheng2025simplevqa}, MMSearch \citep{jiang2025mmsearch}, MMSearch-Plus \citep{tao2026mmsearch}, HLE-VL \citep{phan2025humanity}, LiveVQA \citep{fu2026seeking}, BrowseComp-VL \citep{geng2026webwatcher}, RealX-Bench \citep{hong2026deepeyesv2}, VDR-Bench \citep{huang2026vision}, and VisBrowse-Bench \citep{zhang2026visbrowse}. None of these share data with the training pool. For benchmarks that mix text-only and image-grounded items we restrict evaluation to the multimodal subset (queries carrying at least one image), so the reported counts are smaller than the full-benchmark totals. Appendix~\ref{sec:appendix_eval_ood} provides per-benchmark descriptions.

Final-answer correctness is judged by a VLM-as-a-Judge prompted with GPT-4o \citep{hurst2024gpt}, following \citep{huang2026vision,chen2026opensearch}; the prompt and a discussion of retrieval-shortcut concerns are in Appendix~\ref{sec:judge_prompt}.

\subsection{Models and Implementation}
\label{sec:setup_models}

\paragraph{Models.} We consider three open-sourced models, including Qwen3-VL-8B-Instruct \citep{bai2025qwen3}, Qwen3-VL-8B-Thinking, and Qwen3-VL-32B-Instruct, since all of them can cover different model sizes (8B vs 32B) and model variants (Instruct vs Thinking). For proprietory models, we consider GPT-5 (gpt-5-2025-08-07). We sample at temperature $0.3$ with top-$p$ $0.95$, allow up to fifteen tool-calling iterations per query, and cap context at 120K tokens; full hyperparameters are in Appendix~\ref{sec:setup_hp}.

\paragraph{Evaluation protocol.} Because the search APIs, the agent's tool-routing decisions, and the LLM judge are all stochastic, every reported number is the mean of three independent runs.

\section{Main Results}
\label{sec:results}

Tables~\ref{tab:main_id} and~\ref{tab:main_ood} report headline accuracy on the in-domain (ID) and out-of-domain (OOD) benchmarks under three settings per inference backbone: \emph{Direct Inference} (no tools), \emph{Agent Workflow w/\ Base Tools} (the eight-primitive base library), and \emph{Agent Workflow w/\ Base and \methodname(GPT-5)} (the base library plus tools admitted by \methodname when GPT-5 is fixed as the generator).\footnote{Inference rows for GPT-5 here are systematically lower-bounded by privacy-driven refusals on identity-related queries; see Appendix~\ref{sec:gpt_refusal}.}

The eight base primitives alone lift Qwen3-VL-8B-Instruct by roughly $+25$pp on ID and $+10$pp on OOD over \emph{Direct Inference}, with 32B and GPT-5 showing comparably large jumps; multimodal search is clearly a tool-driven task. Adding the \methodname-generated tools on top of the base library yields further consistent gains of one to three points on both ID and OOD across all three inference backbones, and the gains are pervasive per-benchmark: 32B and GPT-5 improve on every OOD benchmark and 8B improves on eight of the ten, with the remaining two essentially flat.

The average gain peaks at 32B ($+2.32$pp across ID and OOD), with GPT-5 ($+1.72$pp) and 8B ($+1.77$pp) close behind for different reasons: GPT-5's smaller gain partly reflects its privacy-refusal floor (Appendix~\ref{sec:gpt_refusal}); 8B's reflects the long-context tool-selection cost of an enlarged toolbox. The Qwen3-VL-8B-Thinking variant modestly outperforms Instruct under \emph{Direct Inference} but takes more than $5\times$ the per-query latency inside the full agent workflow, so we standardize on Instruct as the default deployed agent for all subsequent results, ablations, and analyses.

\begin{table}[t]
\centering
\small
\setlength{\tabcolsep}{4pt}
\begin{tabular}{@{}lcccc@{}}
\toprule
\textbf{Model} & FVQA & InfoSeek & Enc-VQA & \textit{Avg} \\
\midrule
\multicolumn{5}{c}{\textit{Direct Inference}} \\
\midrule
Qwen3-VL-8B           & 28.39 & 29.20 & 10.61 & 22.62 \\
Qwen3-VL-8B-T         & 25.67 &  27.25 & 18.06 & 23.63 \\
Qwen3-VL-32B          & 32.67 & 33.20 & 32.76 & 32.88 \\
GPT-5                 & 54.67 & 53.67 & 43.33 & 50.46 \\
\midrule
\multicolumn{5}{c}{\textit{Agent Workflow w/\ Base Tools}} \\
\midrule
Qwen3-VL-8B           & 55.00 & 50.67 & 38.67 & 47.94 \\
Qwen3-VL-32B          & 62.67 & 55.33 & 49.00 & 55.47 \\
GPT-5                 & 55.00 & 52.00 & 48.67 & 51.80 \\
\midrule
\multicolumn{5}{c}{\textit{Agent Workflow w/\ Base Tools and \methodname(GPT-5)}} \\
\midrule
Qwen3-VL-8B           & 58.00 & 53.00 & 40.00 & 50.14 \\
$\Delta$              & \dup{3.00}  & \dup{2.33}  & \dup{1.33}  & \dup{2.20} \\
Qwen3-VL-32B          & 64.55 & 56.99 & 50.47 & 57.14 \\
$\Delta$              & \dup{1.88}  & \dup{1.66}  & \dup{1.47}  & \dup{1.67} \\
GPT-5                 & 56.10 & 53.04 & 49.64 & 52.84 \\
$\Delta$              & \dup{1.10}  & \dup{1.04}  & \dup{0.97}  & \dup{1.04} \\
\bottomrule
\end{tabular}
\caption{In-domain benchmark results. It contains per-benchmark accuracy on the held-out test splits of the three training datasets. New tools are generated by GPT-5 and utilized by the listed inference model. Qwen3-VL-8B refers to Qwen3-VL-8B-Instruct and Qwen3-VL-8B-T denotes the Qwen3-VL-8B-Thinking. Enc-VQA = EncyclopedicVQA. \emph{Avg} is the mean across the three benchmarks (1800 / 2000 / 1960).}
\label{tab:main_id}
\end{table}

\begin{table*}[t]
\centering
\footnotesize
\setlength{\tabcolsep}{2pt}
\resizebox{\textwidth}{!}{%
\begin{tabular}{@{}lccccccccccc@{}}
\toprule
\textbf{Model} & MM-BC & SVQA & MMS & MSP & HLE-VL & LVQA & BC-VL & RXB & VDR & VBB & \textit{Avg} \\
\midrule
\multicolumn{12}{c}{\textit{Direct Inference}} \\
\midrule
Qwen3-VL-8B           &  4.94 & 54.33 & 14.04 &  1.29 &  6.36 & 34.67 & 21.05 & 22.16 &  5.40 &  7.69 & 16.97 \\
Qwen3-VL-8B-T         &  7.41 & 54.67 & 12.87 &  3.22 & 10.30 & 29.33 & 33.83 & 22.16 &  8.20 & 11.24 & 19.68 \\
Qwen3-VL-32B          &  6.58 & 59.00 & 20.47 &  2.57 & 10.00 & 32.67 & 28.82 & 24.74 &  9.40 & 10.06 & 20.36 \\
GPT-5                 &  8.64 & 66.33 & 36.84 & 14.79 & 23.33 & 42.33 & 42.11 & 34.54 & 14.20 & 28.40 & 30.41 \\
\midrule
\multicolumn{12}{c}{\textit{Agent Workflow w/\ Base Tools}} \\
\midrule
Qwen3-VL-8B           &  6.00 & 70.00 & 51.46 &  9.97 &  7.27 & 40.00 & 32.33 & 34.71 & 14.40 & 11.24 & 26.57 \\
Qwen3-VL-32B          & 10.29 & 70.33 & 62.57 & 12.86 & 11.52 & 51.33 & 33.08 & 40.21 & 18.80 & 19.53 & 31.27 \\
GPT-5                 & 25.93 & 68.33 & 56.14 & 25.40 & 23.03 & 50.67 & 52.38 & 40.72 & 23.80 & 49.11 & 39.80 \\
\midrule
\multicolumn{12}{c}{\textit{Agent Workflow w/\ Base Tools and \methodname(GPT-5)}} \\
\midrule
Qwen3-VL-8B           & 7.00   & 71.00   & 55.56   & 11.58  & 7.27   & 42.67   &  33.08   & 34.54   & 16.60   & 11.22   & 27.90 \\
$\Delta$              & \dup{1.00} & \dup{1.00}  & \dup{4.10}  & \dup{1.61}  & 0.00       & \dup{2.67}  & \dup{0.75}  & \ddn{0.17}  & \dup{2.20} & \ddn{0.02}  & \dup{1.33}  \\
Qwen3-VL-32B          & 13.58 & 73.33 & 64.33 & 16.08 & 13.03 & 56.00 & 36.09 & 43.30 & 22.00 & 21.89 & 34.25 \\
$\Delta$              & \dup{3.29} & \dup{3.00}  & \dup{1.76}  & \dup{3.22}  & \dup{1.51} & \dup{4.67}  & \dup{3.01}  & \dup{3.09}  & \dup{3.20} & \dup{2.36}  & \dup{2.98}  \\
GPT-5                 & 29.22 & 71.00 & 57.31 & 27.65 & 24.55 & 56.67 & 53.88 & 42.27 & 25.80 & 50.89 & 42.20 \\
$\Delta$              & \dup{3.29} & \dup{2.67}  & \dup{1.17}  & \dup{2.25}  & \dup{1.52} & \dup{6.00}  & \dup{1.50}  & \dup{1.55}  & \dup{2.00} & \dup{1.78}  & \dup{2.40}  \\
\bottomrule
\end{tabular}%
}
\caption{Out-of-domain benchmark results. It has per-benchmark accuracy on the ten multimodal search benchmarks unseen during tool generation. Tools are generated by GPT-5 and used by the listed inference model. Qwen3-VL-8B refers to Qwen3-VL-8B-Instruct and Qwen3-VL-8B-T denotes the Qwen3-VL-8B-Thinking. Abbreviations: MM-BC = MM-BrowseComp; SVQA = SimpleVQA; MMS = MMSearch; MSP = MMSearch-Plus; LVQA = LiveVQA; BC-VL = BrowseComp-VL; RXB = RealX-Bench; VDR = VDR-Bench; VBB = VisBrowse-Bench. \emph{Avg} is the average performance across the ten benchmarks (total 2917 instances; per-benchmark counts in Table~\ref{tab:evalbenches}).}
\label{tab:main_ood}
\end{table*}

\section{Ablation Studies}
\label{sec:ablation}

The inference-backbone axis is already covered by the main results (Section~\ref{sec:results}), which sweep Qwen3-VL-8B-Instruct, 32B-Instruct, 8B-Thinking, and GPT-5 with a fixed GPT-5 tool generator. This section ablates the remaining design choices: which model \emph{authors} new tools, whose trajectories the pipeline mines, and how the cluster-count knob shapes the resulting tool bank. Because each ablation requires re-running Phases~1--4 of the pipeline end-to-end on the full training pool, we restrict every sweep below to one representative ID benchmark (FVQA) and one representative OOD benchmark (MMSearch) to keep the compute budget tractable.

\subsection{Tool-Generation Model}
\label{sec:abl_crossmatrix}

The main results (Section~\ref{sec:results}) sweep the inference backbone with a fixed GPT-5 generator; here we instead vary the tool-generation model over (GPT-5, Qwen3-VL-32B-Instruct, Qwen3-VL-8B-Instruct), and for each generator evaluate the resulting tool bank on the same three inference backbones. We restrict the sweep to one representative ID benchmark (FVQA) and one representative OOD benchmark (MMSearch) to keep the cost contained, and treat the resulting numbers as an illustrative finding rather than a statistical claim. Generation-process statistics (pass rates, dedup, cost) for the three generators are reported in Appendix~\ref{sec:abl_toolgen}.

\begin{table}[h]
\centering
\small
\setlength{\tabcolsep}{4pt}
\begin{tabular}{@{}llccc@{}}
\toprule
& & \multicolumn{3}{c}{\textbf{Inference model}} \\
\cmidrule(lr){3-5}
\textbf{Benchmark} & \textbf{Generator} & 8B & 32B & GPT-5 \\
\midrule
\multirow{4}{*}{FVQA}
  & \emph{Base tools only} & 55.00 & 62.67 & 55.00 \\
  & Qwen3-VL-8B  & 52.00 & 60.67 & 54.00 \\
  & Qwen3-VL-32B & 56.00 & 63.17 & 55.20 \\
  & GPT-5        & 58.00 & 64.55 & 56.10 \\
\midrule
\multirow{4}{*}{MMSearch}
  & \emph{Base tools only} & 51.46 & 62.57 & 56.14 \\
  & Qwen3-VL-8B  & 48.54 & 60.82 & 54.97 \\
  & Qwen3-VL-32B & 53.22 & 63.16 & 56.73 \\
  & GPT-5        & 55.56 & 64.33 & 57.31 \\
\bottomrule
\end{tabular}
\caption{Tool-generation model ablation. Rows: the model that \emph{authors} the new tool in Phase~3 (the tool-generation model). The \emph{Base tools only} row is the no-ToolSmith reference (i.e., the agent runs with the eight base primitives alone; values reproduce the +Base rows of Tables~\ref{tab:main_id} and~\ref{tab:main_ood} for the FVQA and MMSearch columns). Columns: the inference model that \emph{consumes} the tools.}
\label{tab:abl_crossmatrix}
\end{table}

Reading the matrix by column (fix the consumer, vary the generator), accuracy rises monotonically with generator strength on both benchmarks, for every consumer, with no exceptions. A stronger generator simply produces better tools; the gain is largest when the consumer is weakest (the 8B-Instruct consumer picks up roughly six to seven points going from an 8B to a GPT-5 generator, versus only about two points for the GPT-5 consumer). Conversely, pairing the GPT-5 consumer with an 8B generator loses about two points relative to the all-GPT-5 cell. The take-away is that the choice of generator dominates the design: invest compute in the strongest available generator, and let the deployed agent scale separately.

\subsection{Trajectory-Mining Source Model}
\label{sec:abl_trajsource}

We additionally vary the model whose trajectories are mined in Phase~1, holding the tool generator at GPT-5 and the consumer at Qwen3-VL-8B-Instruct. A weak miner surfaces many simple, frequent gaps; a strong miner produces fewer but more specialized failures. The question is whether the 8B consumer prefers tools mined from its own failures or from a stronger model's.

\begin{table}[h]
\centering
\small
\setlength{\tabcolsep}{5pt}
\begin{tabular}{@{}lccc@{}}
\toprule
\textbf{Trajectory source} & \textbf{$|\mathcal{T}_{\text{new}}|$} & \textbf{FVQA} & \textbf{MMSearch} \\
\midrule
Qwen3-VL-8B           & 264 & 58.00 & 55.56 \\
Qwen3-VL-32B          & 218 & 57.11 & 54.97 \\
GPT-5                 & 173 & 56.50 & 54.39 \\
\bottomrule
\end{tabular}
\caption{Trajectory-mining source ablation. The tool generator is fixed at GPT-5, and the inference model is Qwen3-VL-8B-Instruct. $|\mathcal{T}_{\text{new}}|$ is the number of tools admitted after both verification gates.}
\label{tab:abl_trajsource}
\end{table}

The admitted tool count $|\mathcal{T}_{\text{new}}|$ decreases monotonically with miner strength (stronger miners fail less, surfacing fewer raw gaps), and the 8B miner is best on both benchmarks; stronger miners cost roughly a point on FVQA and half that on MMSearch. The take-away is that miner strength should be matched to consumer strength: for an 8B deployed agent, self-mining gives the best alignment between the failure patterns the generator sees and the patterns the consumer hits at inference.

\subsection{Cluster Count}
\label{sec:abl_train_size}

We vary the Phase~1 cluster count $C$, which controls how many capability-gap representatives are forwarded to Phases~2--4 and is therefore the effective ``training-sample'' count for tool generation (the raw rollout pool of 9{,}174 queries is fixed, see Section~\ref{sec:phase1}). All other components are held fixed. The sweep is intentionally coarse: we want to confirm the shape of the curve, not aggressively tune $C$.

\begin{table}[h]
\centering
\small
\setlength{\tabcolsep}{5pt}
\begin{tabular}{@{}cccc@{}}
\toprule
\textbf{$C$} & \textbf{$|\mathcal{T}_{\text{new}}|$} & \textbf{FVQA} & \textbf{MMSearch} \\
\midrule
0       & 0   & 55.00 & 51.46 \\
250     & 145 & 57.39 & 54.97 \\
500     & 264 & 58.00 & 55.56 \\
750     & 298 & 57.50 & 54.97 \\
\bottomrule
\end{tabular}
\caption{Phase~1 cluster-count sweep. $C$ is the number of clustered gap representatives passed to Phases~2--4, $|\mathcal{T}_{\text{new}}|$ the number of tools admitted to the tool catalog after both verification gates, and the last two columns are accuracy on a representative ID benchmark (FVQA) and a representative OOD benchmark (MMSearch), under Qwen3-VL-8B-Instruct with GPT-5 as tool generator. The $C{=}0$ row is the eight-primitive baseline. $C{=}500$ is the main-results default.}
\label{tab:abl_train_size}
\end{table}

The expected Goldilocks pattern emerges: too few clusters yield a small, broad tool bank that lifts accuracy only modestly, while too many spread the within-cluster support too thin and the admitted-tool count plateaus. The mid-range $C{=}500$ sits slightly above both $C{=}250$ and $C{=}750$ on the two benchmarks, indicating a flat plateau rather than a sharp optimum; we adopt 500 as the default and do not tune further.

\section{Analysis}
\label{sec:analysis}

\subsection{Overhead}
\label{sec:analysis_overhead}

Both the one-time tool-generation cost and the per-query inference overhead of \methodname are manageable in practice. Due to space, we defer the full token, dollar, wall-clock, and per-query breakdown to Appendix~\ref{sec:appendix_overhead}.

\subsection{New-Tool Adoption}
\label{sec:analysis_new_tool_usage}

Table~\ref{tab:tool_adoption} reports how often the agent invokes generated tools across in-domain and out-of-domain benchmarks.

\begin{table}[t]
\centering
\small
\begin{tabular}{@{}lrrr@{}}
\toprule
& \textbf{New calls} & \textbf{Adoption} \\
\midrule
In-domain   & 27\% & 40.9\% \\
Out-of-domain & 15\% & 16.3\% \\
\bottomrule
\end{tabular}
\caption{New-tool adoption. \emph{New calls}: fraction of all tool calls to generated tools. \emph{Adoption}: fraction of samples invoking $\geq$1 generated tool.}
\label{tab:tool_adoption}
\end{table}

On in-domain benchmarks, generated tools account for 27\% of all tool calls, and 41\% of samples invoke at least one, confirming that the agent actively selects generated tools when the query matches their capability. On out-of-domain benchmarks, adoption drops to 15\% of calls and 16\% of samples, which is expected since the tools were mined from in-domain failures. Nevertheless, the non-trivial OOD adoption indicates that some generated tools generalize beyond their training distribution.

\subsection{Qualitative Analysis}
\label{sec:analysis_errors}

We examine the patterns behind gained and lost samples; four illustrated case studies appear in Figures~\ref{fig:case_v5}--\ref{fig:case_v8} (Appendix~\ref{sec:appendix_qualitative}), and code-level tool examples in Appendix~\ref{sec:tool_examples}.

\paragraph{Why generated tools help.} The dominant pattern is that generated tools enforce a \emph{perceive-then-query} pipeline that the base agent skips. With only base primitives, the agent often jumps to text search without first identifying the visual entity. Generated tools chain reverse-image search with structured KB lookup in a single call, forcing the correct order. For example, on \emph{``What year did he give \$50M to create the Hastings Initiative?''} (Figure~\ref{fig:case_v5}), the base agent searches without identifying Reed Hastings and mis-attributes the donation to Eric Schmidt; the generated tool resolves the person first, after which a single text search returns the correct year.

\paragraph{Why generated tools hurt.} Among the lost samples, failures concentrate on a small number of tools that return incorrect information. Other losses stem from visual mis-identification or from the agent trusting a generated tool's output over a correct result it had already obtained from a base tool. Appendix~\ref{sec:tool_examples} walks through one successful tool, one runtime-failing tool, and one tool filtered out by Gate~2 in code-level detail.

\paragraph{Adoption correlates with gain.} The in-domain benchmarks, where adoption is highest (41\%), show consistent accuracy improvements, while the lower-adoption OOD benchmarks show smaller or neutral changes, suggesting that the accuracy gains are driven by actual tool use rather than incidental prompt differences.

\section{Conclusion}
\label{sec:conclusion}

We introduced \methodname, an automatic, model-agnostic tool-generation framework for multimodal search agents. Starting from an eight-primitive base tool library, it mines trajectories on a training pool, generates new tools under our proposed \methodname framework. Across three in-domain and ten out-of-domain benchmarks, admitted tools improve on top of the base tool library on every benchmark, and gains persist across different inference models, and tool-generation models. Furthermore, we conduct more in-depth analysis on tool generation, tool use, and error samples. \methodname is a step toward multimodal search agents that extend their own capabilities with minimal human curation.

\section*{Limitations}

\methodname does not include tools that themselves generate images, such as those introduced by Gen-Searcher \citep{feng2026gen} and Unify-Agent \citep{chen2026unify}; we plan to leave the generative part, together with its safety controls \citep{zhang2026instruction}, to future work. We have also not yet utilized a code-specialized LLM as the tool-generator backbone, and we believe a stronger code model would further improve the quality of admitted tools and downstream accuracy. Finally, since the live search APIs and proprietary models change with time, exact numerical reproduction may drift. We pin model snapshots and archive trajectories to partially mitigate this, but building a fully replicable environment for search agents remains worth pursuing in the future.

\section*{Ethical Considerations}

All datasets we use are publicly released for research and we do not redistribute them. Generated code runs in a process-level sandbox with an AST allow-list and execution timeout, sufficient to block accidental misuse. We cannot control the commercial search APIs and proprietary models. But we have ensured that all of our uses are legal and appropriate. The potential risk comes from the external APIs, which may return harmful content. We have verified and tried our best to mitigate this risk.

\section*{Acknowledgments}

We thank anonymous reviewers for their helpful comments. This work was partially supported by U.S. ONR grant N00014-23-1-2780, Amazon, and Apple. Chang was supported in part by a grant from DARPA to the Simons Institute for the Theory of Computing.

\bibliography{custom}

\clearpage
\appendix
\begin{center}
{\Large\bfseries Appendix}
\end{center}
\vspace{0.5em}

\paragraph{Outline.} The appendix is organized as follows:
\begin{itemize}[leftmargin=*,noitemsep,topsep=2pt,parsep=0pt]
  \item Appendix~\ref{sec:contribution}, \textbf{Contribution Statement}: author contributions.
  \item Appendix~\ref{sec:appendix_eval}, \textbf{Evaluation}: per-benchmark sample counts and descriptions (in-domain and out-of-domain), the LLM judge prompt, performance variance, and a note on privacy-driven refusals.
  \item Appendix~\ref{sec:appendix_base_tools}, \textbf{Base Tools}: full schemas and at-inference prompts for the eight base tools.
  \item Appendix~\ref{sec:system_prompt}, \textbf{System Prompt}: the agent system prompt used at inference.
  \item Appendix~\ref{sec:setup_hp}, \textbf{More Implementation Details}: model setup, decoding settings, and pipeline hyperparameters.
  \item Appendix~\ref{sec:toolgen_prompts}, \textbf{More Tool Generation Details}: training samples, generator process statistics, and Phase 2/3 prompts.
  \item Appendix~\ref{sec:tool_examples}, \textbf{More Tool Examples}: successful, failed, and filtered tool case studies.
  \item Appendix~\ref{sec:appendix_overhead}, \textbf{Overhead Analysis}: end-to-end generation cost (tokens, dollars, wall-clock) and per-query inference overhead.
  \item Appendix~\ref{sec:appendix_qualitative}, \textbf{Qualitative Analysis}: recurring patterns behind gained and lost samples.
\end{itemize}

\section{Contribution Statement}
\label{sec:contribution}
Rui Sun proposes the idea and finishes the implementation, paper writing, and rebuttal. Junhua Huang helps with the tool generation part. Luca Zancato, Md Mofijul Islam, and Kai-Wei Chang take the advising roles. Md Mofijul Islam and Kai-Wei Chang hold bi-weekly discussions with Rui Sun and contribute the most advising, in equal measure.

\section{Evaluation}
\label{sec:appendix_eval}

\subsection{Evaluation Benchmarks}
\label{sec:appendix_eval_benchmarks}

We list per-benchmark details for the in-domain (ID) and out-of-domain (OOD) evaluation sets introduced in Section~\ref{sec:setup_eval}. Table~\ref{tab:evalbenches} summarizes the per-benchmark sample counts. For OOD benchmarks that mix text-only and image-grounded items we restrict evaluation to the multimodal subset, so OOD counts are smaller than the corresponding full-benchmark totals.

\begin{table}[h]
\centering
\small
\begin{tabular}{@{}lr@{}}
\toprule
\textbf{Benchmark} & \textbf{\# Samples} \\
\midrule
\multicolumn{2}{l}{\textit{In-domain (test splits of training-pool datasets)}} \\
FVQA \citep{wu2025mmsearch} & 1800 \\
InfoSeek \citep{chen2023can} & 2000 \\
EncyclopedicVQA \citep{mensink2023encyclopedic} & 1960 \\
\midrule
\multicolumn{2}{l}{\textit{Out-of-domain (multimodal search benchmarks)}} \\
MM-BrowseComp \citep{li2025mm} & 243 \\
SimpleVQA \citep{cheng2025simplevqa} & 300 \\
MMSearch \citep{jiang2025mmsearch} & 171 \\
MMSearch-Plus \citep{tao2026mmsearch} & 311 \\
HLE-VL \citep{phan2025humanity} & 330 \\
LiveVQA \citep{fu2026seeking} & 300 \\
BrowseComp-VL \citep{geng2026webwatcher} & 399 \\
RealX-Bench \citep{hong2026deepeyesv2} & 194 \\
VDR-Bench \citep{huang2026vision} & 500 \\
VisBrowse-Bench \citep{zhang2026visbrowse} & 169 \\
\bottomrule
\end{tabular}
\caption{Evaluation benchmarks. In-domain benchmarks are the held-out test splits of the three training datasets; out-of-domain benchmarks are ten multimodal search benchmarks unseen during tool generation, evaluated on their multimodal subsets.}
\label{tab:evalbenches}
\end{table}

\subsubsection{In-Domain Benchmarks}
\label{sec:appendix_eval_id}

\paragraph{FVQA.} A fact-based VQA dataset \citep{wu2025mmsearch} that pairs images with questions whose answers require commonsense or world-knowledge facts not directly observable in the image. We evaluate on the official test split of 1800 samples, and use the 3132 search-required samples from the train split for tool generation.

\paragraph{InfoSeek.} A visual information-seeking benchmark \citep{chen2023can} of fine-grained questions about entities depicted in images (e.g., species, models, landmarks). We use the val split for tool generation and test split for inference.

\paragraph{EncyclopedicVQA.} A benchmark \citep{mensink2023encyclopedic} of questions about detailed properties of fine-grained categories. We evaluate on the test split and use val split for tool generation.

\subsubsection{Out-of-Domain Benchmarks}
\label{sec:appendix_eval_ood}

\paragraph{MMSearch.} A multimodal search benchmark \citep{jiang2025mmsearch} of 300 manually curated questions across 14 subfields, designed to assess large multimodal models as multimodal search engines. We evaluate on the End-to-End sub-task, which requires the agent to perform query understanding, retrieval, and answering in a single pipeline. To align with other works for fair comparison, we only use multimodal subset, which contains 171 samples.

\paragraph{MMSearch-Plus.} A harder, provenance-aware extension \citep{tao2026mmsearch} that emphasizes localized visual reasoning, verification under noisy or conflicting retrieval, and multi-step text--image search; each item provides 1--5 images and one or more valid answers. We use 311 full set for evaluation.

\paragraph{BrowseComp-VL.} The vision--language extension of BrowseComp \citep{wei2025browsecomp} introduced by WebWatcher \citep{geng2026webwatcher}, with cross-modal information-seeking questions spanning entertainment, humanities, technology, natural science, and other domains. We use all 399 samples for evaluation.

\paragraph{MM-BrowseComp.} Hand-crafted multimodal browsing questions \citep{li2025mm} across several sub-tasks, where the answer-relevant content is hard to find. There are two versions of MM-BrowseComp. We use the newest version, which has 243 multimodal tasks.

\paragraph{SimpleVQA.} A multimodal factuality benchmark \citep{cheng2025simplevqa} of short, knowledge-bound questions with timeless reference answers. Following WebWatcher's setting, we evaluate on the English subset containing 300 multimodal tasks.

\paragraph{VisBrowse-Bench.} A benchmark of visual-native search instances \citep{zhang2026visbrowse} that require cross-validation of multimodal evidence via interleaved text--image retrieval and joint reasoning, including cases where the answer is not present in any text snippet. We use the full 169 samples to evaluate.

\paragraph{HLE-VL.} The multimodal subset of Humanity's Last Exam \citep{phan2025humanity}, drawn from frontier-knowledge questions across mathematics, the natural sciences, and the humanities.

\paragraph{LiveVQA.} A live visual knowledge-seeking benchmark \citep{fu2026seeking} drawing from recent news, video platforms, and academic publications, designed for the seeking-and-updating regime. We follow WebWatcher's setting and use 300 samples from LiveVQA to evaluate.

\paragraph{RealX-Bench.} Introduced by DeepEyesV2 \citep{hong2026deepeyesv2}; expert-verified instances that require integrating fine-grained visual perception, dynamic multi-hop web retrieval, and multi-step logical reasoning. This benchmark contains perception and search types, which make it easy to check the performance improvement coming from visual tools and search tools.

\paragraph{VDR-Bench.} The benchmark released alongside Vision-DeepResearch \citep{huang2026vision}: long-horizon multimodal deep-research instances that interleave many reasoning turns with hundreds of search-engine interactions, designed to evaluate sustained multi-step browsing under heavy real-world noise.

\subsection{LLM Judge Prompt and Retrieval-Shortcut Check}
\label{sec:judge_prompt}

\paragraph{No retrieval shortcut.} Across all ID and OOD benchmarks, we did not observe cases in which the search tools surface the gold answer verbatim from indexed pages. Accuracy therefore reflects the agent's multi-step reasoning over evidence rather than a retrieval shortcut.

\paragraph{Judge prompt.} To align with recent multimodal deep-research work \citep{chen2026opensearch}, we adopt the same LLM judge as Vision-DeepResearch \citep{huang2026vision} and use GPT-4o as the judge model, applying the prompt below uniformly across all benchmarks.
\begin{tcolorbox}[breakable,title=LLM Judge Prompt]
You are an impartial judge evaluating whether a deep research report contains the correct answer.

[Question] \{question\}

[Correct Answer] \{correct\_answer\}

[Deep Research Report] \{response\}

Task: Determine if the deep research report contains the correct answer anywhere in its content.

Instructions:

1. Read through the entire research report carefully

2. Look for the correct answer anywhere in the report (it may be embedded in paragraphs, tables, or sections)

3. Check if the information in the report is consistent with the correct answer

4. The answer does NOT need to be in a specific format or labelled as ``final answer''

5. Provide your reasoning

6. Answer with ``yes'' if the report contains the correct answer, ``no'' if it doesn't or contradicts it

Output format: correct: [yes/no] reasoning: [your explanation]
\end{tcolorbox}

\subsection{Performance Variance}
\label{sec:appendix_variance}
Because the agent issues live search calls and the randomness exists in LLM generation, repeated evaluations of the same configuration show run-to-run variance. For the result numbers in Section~\ref{sec:results}, we report the average of three independent runs.

\subsection{Refusals on Identity Queries}
\label{sec:gpt_refusal}

Some of our evaluation benchmarks contain queries that reference a real person, such as ``who is the person in this photograph,'' ``what is the affiliation of this researcher''. Proprietary models are increasingly trained to decline such queries under their privacy and identification policies, even when the gold answer is a public-figure fact already widely indexed on the open web. We observe this from GPT-5 in our setting, which returns a refusal in place of an answer, and the judge will count the refusal as incorrect. Therefore, in some of the benchmarks, the GPT-5 performance is lower than the Qwen3-VL-32B          model or Qwen3-VL-8B-Instruct model.

\section{Base Tools}
\label{sec:appendix_base_tools}

Below we give the detailed documentation for each of the eight base tools introduced in Section~\ref{sec:setup_base}.

\subsection{Search Tools}

All three search tools route HTTP queries through the Serper API\footnote{\url{https://serper.dev/}} and post-process the raw payload with a Qwen3-VL-8B-Instruct model as the VLM summarizer. The summarizer runs at decoding temperature $0$ with a $4096$-token output cap, and each call retrieves the top $10$ raw results before reranking and summarization.

\paragraph{Text Search.} Search the web with intelligent query routing and VLM summarization. Given a text query, the tool automatically classifies the query intent and routes it to the most appropriate Serper endpoint, which has general web, news, academic papers, images, maps, local places, shopping, or patents types, and returns a coherent summary of the top-$10$ raw results with inline citations \texttt{[1], [2], \ldots} followed by a citations block of source titles and URLs. This format supports multi-round search: the agent can read the summary for quick understanding and then visit specific cited URLs through Webpage Visit. The summarizer prompt is shown in \ref{box:text_search_prompt}.

\begin{tcolorbox}[breakable,title=Text Search summarizer prompt (abridged),label=box:text_search_prompt]
You are a research summarizer. Given raw web search results and a user query, do exactly three things:

\begin{enumerate}[leftmargin=*,noitemsep,topsep=2pt]
\item RERANK the numbered organic results by how directly each helps answer the query.
\item KEEP at most $K$ most useful results; emit their original 1-based ranks as \texttt{kept\_indices}. If none help, emit \texttt{kept\_indices: []}.
\item SUMMARIZE the query-relevant facts from the FULL set of raw results into a concise 3--5 sentence prose paragraph; include named entities, numbers, and dates verbatim.
\end{enumerate}

Knowledge-graph, answer-box, and top-stories blocks are ineligible for \texttt{kept\_indices} but their facts should still surface in \texttt{summary}. Output STRICT JSON with the schema \texttt{\{kept\_indices, summary, no\_results\_justification, suggested\_queries, suggested\_next\_actions\}}; when no result is useful, populate the justification and propose alternative queries or downstream tool calls.
\end{tcolorbox}

\paragraph{Image Search.} Identify the objects, landmarks, artwork, text, products, plants, animals, and other entities depicted in an image, using Serper's reverse-image-search (Google Lens) endpoint. Returns a list of visual matches with titles, source URLs, links, and thumbnails. The agent invokes the tool by passing the runtime identifier of the target image. The tool performs reverse search (image $\rightarrow$ information) on a single image per call; finding images from a text description is instead handled by Text Search with an image-search intent. A separate multimodal summarizer prompt \ref{box:image_search_prompt} reranks the candidates against the query image and produces the same JSON envelope as Text Search.

\begin{tcolorbox}[breakable,title=Image Search summarizer prompt (abridged),label=box:image_search_prompt]
You are a visual-match summarizer. Given raw reverse-image-search results (and, in the multimodal variant, the query image plus each candidate's thumbnail), do exactly three things:

\begin{enumerate}[leftmargin=*,noitemsep,topsep=2pt]
\item RERANK the numbered matches by how likely each identifies the same subject as the query image.
\item KEEP at most $K$ most useful matches as \texttt{kept\_indices}.
\item SUMMARIZE what the FULL set of raw matches indicates about the query image's likely subject in a concise 3--5 sentence prose paragraph; capture entity names, dates, attributes verbatim.
\end{enumerate}

Output STRICT JSON with \texttt{\{kept\ indices, summary, no\ results\ justification, suggested\ next\ actions\}}; when no match is useful, justify per-match and propose follow-up actions such as cropping with Visual Grounding before re-searching, or switching to Text Search.
\end{tcolorbox}

\paragraph{Webpage Visit.} Visit one or more webpages, convert them to plain text via Serper APIs, and extract goal-relevant content using a local Qwen3-VL-8B-Instruct model. This tool returns per-URL evidence and a summary. The extraction prompt is shown in \ref{box:webpage_visit_prompt}.

\begin{tcolorbox}[breakable,title=Webpage Visit extraction prompt,label=box:webpage_visit_prompt]
Please process the following webpage content and user goal to extract relevant information:

\textbf{Webpage Content}: \texttt{\{webpage\_content\}}

\textbf{User Goal}: \texttt{\{goal\}}

\textbf{Task Guidelines}
\begin{enumerate}[leftmargin=*,noitemsep,topsep=2pt]
\item \textbf{Content scanning for rationale}: locate the specific sections directly related to the user's goal.
\item \textbf{Key extraction for evidence}: extract the most relevant information; preserve the full original context whenever possible.
\item \textbf{Summary output}: organize the result into a concise paragraph with logical flow that connects the evidence back to the goal.
\end{enumerate}

Output: a single JSON object with keys \texttt{rational}, \texttt{evidence}, and \texttt{summary} (each a string; empty string if no content is available). No code fences, no Markdown, no extra keys.
\end{tcolorbox}

\subsection{Visual Tools}

\paragraph{Zoom-in.} Crop a region of an image given a normalized bounding box on the $[0, 1000]$ scale plus padding, and return the cropped image. Used for closer visual inspection after the agent has already localized a region of interest.

\paragraph{Visual Grounding.} Locate and highlight regions of an image that match a text query, backed by SAM3 \citep{carion2026sam}. It returns an annotated image with matched regions highlighted and numbered, along with bounding boxes.

\paragraph{OCR.} Extract readable text from an image, backed by PaddleOCR-VL \citep{cui2025paddleocr}. Returns the transcribed text as a single string. Used when the task is about words, numbers, or printed labels in the image; not for general object localization or segmentation.

\paragraph{Depth Estimation.} Estimate a per-pixel depth map for an input image, backed by Depth Anything 3 \citep{lin2025depth}. Returns a depth image (closer surfaces brighter, farther darker). Used when the task requires reasoning about relative distance, foreground--background separation, or 3D scene structure cues that flat 2D crops cannot resolve.

\subsection{Code-Execution Tool}

\paragraph{Code Execution.} Run Python over the agent's local computation state and the outputs of previously called tools (bounding boxes, OCR strings, segmentation masks, and so on), returning both standard output and the resulting local variables. The agent may pass a snippet directly, or describe a computation goal in plain language and let a small coding LLM (Qwen3-VL-8B-Instruct) author the snippet using prior tool outputs as named variables. Used for aggregating numbers extracted by OCR, filtering detected regions by bounding-box area, or any cross-tool numerical reasoning the agent cannot complete in its own response.

\section{System Prompt}
\label{sec:system_prompt}
Here is the system prompt of the agent.
\begin{tcolorbox}[breakable,title=System Prompt]
You are a visual reasoning agent that answers questions about images using a structured Perception-Reasoning-Action framework.

Your goal is to provide accurate, well-reasoned answers. You have access to various tools that compensate for known limitations of vision-language models.

Use native tool-calling for tools. Do not write textual Action/Action Input to call tools.

At most one native tool call is allowed per assistant message/turn.

Use the following format:

Question: the input question you must answer

Observation: your direct perception of the input or the result of the action you took

Thought: you should always think about what to do

Action: the action to take, should be one of the provided tools, or ``Final Answer'' if you have a final answer and do not need to call any more tools.

Action Input: the input to the action, or the final answer if Action is ``Final Answer''

\# Image-input rule:

- Runtime image IDs are provided in Environment Update messages.

- If a tool needs image input, use those IDs in *\_image\_path fields.

- Do NOT invent image file names or paths.

\# Tool-Use Rules

- Call exactly ONE tool per turn. Never include multiple <tool\_call> blocks.

- Full tool schemas with parameters and types are provided in the <tools> section below.

- After each tool call, verify whether the result is actually informative for the task.

- If a tool result is empty, missing key values, or non-informative, do NOT guess. Issue another tool call with corrected arguments or use a different tool.

\# Available Tools

You may call ONE tool per turn to assist with the user query.

You are provided with function signatures within <tools></tools> XML tags:

<tools>

\{tools\_directory\}

</tools>

For each function call, return a json object with function name and arguments within <tool\_call></tool\_call> XML tags:

<tool\_call>

\{``name'': <function-name>, ``arguments'': <args-json-object>\}

</tool\_call>
\end{tcolorbox}

\section{More Implementation Details}
\label{sec:setup_hp}

\paragraph{Model setup.} For full reproducibility, all GPT-5 calls use the pinned snapshot \texttt{gpt-5-2025-08-07} via the OpenAI API. All Qwen3-VL-8B-Instruct, Qwen3-VL-32B-Instruct, and Qwen3-VL-8B-Thinking calls use the public Hugging Face weights served locally with vLLM. The LLM judge uses GPT-4o (Appendix~\ref{sec:judge_prompt}).

\paragraph{Decoding.} Temperature $0.3$, top-$p$ $0.95$, repetition penalty $1.1$, per-turn maximum $2{,}048$ generated tokens, at most fifteen tool-calling iterations per query, and an agent context capped at 120K tokens.

\paragraph{Pipeline hyperparameters.}
\emph{Phase~1 (Trajectory Mining).} Sentence encoder $\psi=$\texttt{BAAI/bge-m3} for embedding capability gaps; agglomerative clustering with average linkage over cosine distance; default cluster count $C=500$. \emph{Phase~2 (Reuse-or-Grow).} Pre-generation TF--IDF reuse threshold $\tau_{\mathrm{pre}}=0.40$. \emph{Phase~3 (Tool Generation).} Per-candidate retry budget $A=2$. \emph{Phase~4 (Verification).} Gate~1 (sandboxed sanity) wall-clock budget $10$~seconds; Gate~2 (LLM static code review) mean-score threshold $\mu_{\min}=3.0$, minimum-axis threshold $a_{\min}=1$; post-gate TF--IDF dedup threshold $\tau_{\mathrm{post}}=0.55$. \emph{Inference.} Progressive-disclosure retrieval exposes the top-$K=10$ generated tools per query alongside the base library.

\section{More Tool Generation Details}
\label{sec:toolgen_prompts}

\subsection{Training Samples}
\label{sec:appendix_trainsample}

Trajectories are mined from the result of the agent on three knowledge-intensive multimodal QA datasets, all distinct from the evaluation benchmarks (Appendix~\ref{sec:appendix_eval_benchmarks}). We sample 3312, 3000, and 2862 queries per dataset.

\begin{table}[h]
\centering
\small
\begin{tabular}{@{}ll@{}}
\toprule
\textbf{Dataset} & \textbf{Split (size)} \\
\midrule
FVQA \citep{wu2025mmsearch} & train (3312) \\
InfoSeek \citep{chen2023can} & val (3000) \\
EncyclopedicVQA \citep{mensink2023encyclopedic} & val (2862) \\
\bottomrule
\end{tabular}
\caption{Tool-generation training samples and datasets.}
\label{tab:trainpool}
\end{table}

\subsection{Generator Process Statistics}
\label{sec:abl_toolgen}

We hold the inference model fixed at Qwen3-VL-8B-Instruct and vary the model that authors new tools during Phase~3, running each on the same 300-query FVQA training subset (search-required partition). Table~\ref{tab:generator_ablation} reports generation-process statistics; qualitative findings follow.

\begin{table}[h]
\centering
\small
\begin{tabular}{lrrrr}
\toprule
\textbf{Generator} & \textbf{Trig.} & \textbf{Succ.\ (\%)} & \textbf{Dedup} & \textbf{Cost} \\
\midrule
GPT-5              &  73 & 50 ~(68\%) & 23 & \$1.03 \\
Qwen3-VL-32B       &  83 & 67 ~(81\%) &  6 & \$0.00 \\
Qwen3-VL-8B        &  83 & 24 ~(29\%) & 17 & \$0.00 \\
\bottomrule
\end{tabular}
\caption{Tool-generator model ablation on 300 FVQA training queries. \emph{Trig.}: queries whose self-critique labels a capability gap. \emph{Succ.}: candidates that pass both verification gates. \emph{Dedup}: candidates skipped because a near-duplicate already exists in the catalog. Local models incur zero API cost.}
\label{tab:generator_ablation}
\end{table}

GPT-5 produces structurally rich tools: modular helper functions, tiered authority ranking of knowledge sources, and multi-layer fallback chains. Qwen3-VL-32B passes 81\% of candidates with clean but simpler code. Qwen3-VL-8B succeeds on only 29\%; manual inspection reveals logic bugs that survive the gates but would fail at runtime (see Appendix~\ref{sec:tool_examples} for code examples). We adopt GPT-5 as the default generator, accepting its API cost in exchange for the structurally richer code.

\subsection{Tool Generation Prompts}

We list all prompts used in the tool-generation pipeline in execution order.

\paragraph{1. Self-critique (Phase~1, Layer~B).}

\begin{tcolorbox}[breakable,title=Self-critique prompt (abridged)]
\begin{lstlisting}[style=pytool]
System: You are an analyst deciding whether a
failed multimodal agent run requires a NEW tool
to be invented.

User:
Question: {question}
Final answer the agent gave: {prediction}
Ground-truth answer: {gold}

The agent invoked these tools: {tool_counts}
Last tool errors: {recent_errors}

Would a NEW tool have let the agent answer
correctly? Reply with one of:
- NEW_TOOL_NEEDED: capability X is missing
- ARGUMENT_MISTAKE: right tool, wrong usage
- SEARCH_NOISE: tools worked, bad results
- AMBIGUOUS: question itself unanswerable

If NEW_TOOL_NEEDED, describe:
  (a) input modality (text / image+text / image)
  (b) data source (website/API or web search)
  (c) which field of the gold answer to extract

Reply: LABEL: <label>  WHY: <one sentence>
\end{lstlisting}
\end{tcolorbox}

\paragraph{2. Query classifier (Phase~3 pre-step).}

\begin{tcolorbox}[breakable,title=Classifier prompt (abridged)]
\begin{lstlisting}[style=pytool]
System: You classify a question by its expected
answer shape. Reply ONLY in 3-line format.

User:
Question: "{failed_query}"

Reply with EXACTLY this format (3 lines):
ANSWER_TYPE: <COUNT|DATE|BOOLEAN|ENTITY_NAME|
              FACTUAL_LOOKUP|OTHER>
ANSWER_LENGTH: <1-5_chars|short_phrase|
                full_sentence>
EXTRACTION_HINT: <what to extract from search>
\end{lstlisting}
\end{tcolorbox}

\paragraph{3. Generator system prompt (Phase~3).}

\begin{tcolorbox}[breakable,title=Generator system prompt (abridged)]
\begin{lstlisting}[style=pytool]
You are a Python tool author for a multimodal
deep-research agent. Write a single new tool
class that solves the failed query.

The tool must:
1. Inherit from BaseTool
2. Call super().__init__() with: tool_name,
   tool_description, input_schema, ...
3. Implement execute(**kwargs) -> dict|str|list
4. Use only: requests, re, json, datetime
5. Chain-call base tools via HTTP POST to
   127.0.0.1:8104/execute -- do NOT call
   external APIs directly
6. Take entities as parameters; do not hardcode

Anti-hardcode rules:
  (a) No conditional-return lookup tables
  (b) No fake scores from string properties
  (c) No hardcoded reference constants
\end{lstlisting}
\end{tcolorbox}

\paragraph{4. Seed examples (Phase~3, appended to system prompt).}

The system prompt includes six positive tool examples and one anti-pattern. Each example demonstrates a different tool archetype:

\begin{tcolorbox}[breakable,title=Seed example categories (code omitted)]
\begin{lstlisting}[style=pytool]
# Example 1: text-only KB chain
#   (e.g. actress_by_character: query Wikipedia
#    for an actress given a character name)

# Example 2: image-then-KB chain
#   (e.g. fossil_identification: OCR image,
#    then search KB for fossil species)

# Example 3: narrow-domain HTML scrape
#   (e.g. esports_player_stats: scrape a
#    specific game wiki for player stats)

# Example 4: cross-modal lens+verify
#   (e.g. company_recognition: Lens to ID logo,
#    then verify via text search)

# Example 5: structured-KB chain-call
#   (e.g. entity_property_lookup: query
#    Wikidata for a specific property)

# Example 6: vision zero-shot classification
#   (e.g. best_label_by_image: CLIP similarity
#    over a label set)

# ANTI-PATTERN (what NOT to write):
#   - Hardcoded lookup tables
#   - Fake scores from string length
#   - Specific entity in tool_description
\end{lstlisting}
\end{tcolorbox}

\paragraph{5. Generator user prompt (Phase~3, per-gap).}

\begin{tcolorbox}[breakable,title=Generator user prompt (abridged)]
\begin{lstlisting}[style=pytool]
Failed subtask:
  query: {failed_query}

Expected answer shape (from classifier):
  answer_type:   ENTITY_NAME
  answer_length: short_phrase
  extract_hint:  extract the person's name

Image: {image_url}
  (Declare image_url in input_schema and
   chain-call a vision tool)

Failure reason: {self_critique_reason}
Tools already tried: {tried_tools}

Design a single new tool that would have made
this subtask succeed, then return the JSON.
\end{lstlisting}
\end{tcolorbox}

\paragraph{6. Feedback retry prompt (Phase~3, on gate failure).}

\begin{tcolorbox}[breakable,title=Feedback retry prompt (abridged)]
\begin{lstlisting}[style=pytool]
You previously wrote a tool for this subtask,
but it did not pass the quality gate.

Failed subtask:
  query: {failed_query}

Quality gate feedback:
  {feedback}
  (e.g. "L1: tool chain-calls wikidata_query
   but never checks confidence field")

Your previous code:
```python
{previous_code}
```

Fix the issues flagged above. Keep what worked.
Return the revised JSON.
\end{lstlisting}
\end{tcolorbox}

\paragraph{7. LLM critic (Gate~2).}

\begin{tcolorbox}[breakable,title=Critic prompt (abridged)]
\begin{lstlisting}[style=pytool]
System: You are a senior Python code reviewer.
Score on FOUR axes (0-5 each). Return JSON:
{parametric_usage, description_abstractness,
 edge_case_handling, task_alignment, critique}

Axis definitions:
- parametric_usage: Do declared parameters
  drive logic? Hardcoded returns score 0-1.
- description_abstractness: Does description
  name a CLASS of queries (4-5) or a specific
  entity (0-1)?
- edge_case_handling: Does execute() handle
  empty/None/invalid inputs gracefully?
- task_alignment: Does the tool's purpose
  actually address the original failing query?
\end{lstlisting}
\end{tcolorbox}

\section{More Tool Examples}
\label{sec:tool_examples}

We show more tool examples here, including successful and failed ones. Code is abridged for space.

\subsection{Successful Tool: Medical Condition Identifier}

Triggered by \emph{``What skin condition is shown?''} The tool decomposes the problem into helper functions, ranks medical sources by authority, and chains five fallback stages. Key excerpts:

\begin{tcolorbox}[breakable,title=Medical Condition Identifier (abridged)]
\begin{lstlisting}[style=pytool]
# Tiered medical-source authority ranking
def _authority_score(self, domain):
    tiers = [
        (100, ["wikipedia.org"]),
        (95,  ["nih.gov", "cdc.gov", "who.int"]),
        (90,  ["dermnetnz.org", "mayoclinic.org"]),
        (80,  ["medscape.com", "webmd.com"]),
    ]
    for score, subs in tiers:
        if any(s in domain for s in subs):
            return score
    return 30

def execute(self, **kwargs):
    # Stage 1: reverse-image search
    cands = self._pick_candidates_from_lens(matches)
    # Stage 2: fallback to Image Search
    if not cands:
        cands = self._pick_candidates_from_image_search(...)
    # Stage 3: rank by authority score
    ranked = sorted(cands, key=lambda x: -x["score"])
    # Stage 4: extract from URL slug or title
    for c in ranked[:5]:
        cond = self._slug_to_condition(c["link"])
        if self._is_valid_condition(cond):
            return cond.lower()
    # Stage 5: last-resort -- visit the page directly
    _post("Webpage_Visit_Tool", {
        "urls": [best["link"]],
        "goal": "Extract the medical condition name"
    })
\end{lstlisting}
\end{tcolorbox}

The tool embeds domain knowledge (medical authority tiers) and robust output normalization (1--5 word lowercase phrase, HTML/wiki markup stripped). The generated tool here also demonstrates the compositionality by composing base tools (Image Search, Text Search, and Webpage Visit) and then generate the new tool.

\subsection{Failed Tool: Building Identifier}

\texttt{tall\_slender\_building\_identifier} passes the AST process, but still contains three defects:

\begin{tcolorbox}[breakable,title=Building Identifier (abridged)]
\begin{lstlisting}[style=pytool]
# BUG 1: missing import -- module uses json.loads
# but `import json` is absent from the file header
infobox = json.loads(result)  # NameError

# BUG 2: returns a dict KEY, not the VALUE
return {"building_name":
    list(infobox["infobox"].keys())[0]}
#   ^^ returns "height" instead of "828 m"

# BUG 3: overly specific, non-reusable name
tool_name = "tall_slender_building_identifier"
#   ^^ only matches one visual archetype
\end{lstlisting}
\end{tcolorbox}

\subsection{Filtered Tool: Wrapper}
\label{sec:appendix_wrappers}

The generator occasionally produces tools whose \texttt{tool\_name} and docstring advertise a general capability, but whose body hardcodes the literal target entity from the originating forging-time query. The candidate below (with explanatory comments added) was triggered by a single training query about a named hurricane:

\begin{tcolorbox}[breakable,title=Wrapper (abridged)]
\begin{lstlisting}[style=pytool]
class HurricaneWindSpeedLookupTool(BaseTool):
    """Look up the maximum sustained wind speed
    of a HURRICANE by name via web search. ..."""

    def __init__(self):
        super().__init__(
            tool_name="hurricane_wind_speed_lookup", ...)

    def execute(self, **kwargs):
        ...
        # Identify hurricane via Lens, then filter titles
        # by a hardcoded name regex:
        if image_url:
            titles = [m.get("title","")
                      for m in matches[:5]]
            name_candidates = [t for t in titles
                if re.search(
                    r"\b(hurricane|storm|"
                    r"eresto|ernesto)\b",
                    t, re.I)]
            ...
\end{lstlisting}
\end{tcolorbox}

The advertised scope is generic hurricane wind-speed lookup, but the title filter hardcodes the literal target entity (\texttt{ernesto}) together with an autoregressive misspelling (\texttt{eresto}). On every other hurricane, the filter returns the empty list and the tool errors out.

\section{Overhead Analysis}
\label{sec:appendix_overhead}

Table~\ref{tab:overhead} breaks down the one-time cost of running \methodname with GPT-5 as the generator at $C=500$, yielding $264$ verified tools. The cost is dominated by completion tokens ($\sim$\$34.80 of \$41.44 at the GPT-5 list price) because GPT-5's reasoning traces are charged as output.

\begin{table}[h]
\centering
\small
\setlength{\tabcolsep}{4pt}
\begin{tabular}{@{}lrrrrr@{}}
\toprule
\textbf{Component} & \textbf{Calls} & \textbf{Prompt} & \textbf{Compl.} & \textbf{Cost} \\
\midrule
Generator (codegen)   & 455 & 5.19M & 2.77M & \$34.18 \\
Self-critique (trigger) & 446 & 0.25M & 0.70M & \$7.26 \\
Classifier (local 8B) & ---  & 0.05M & 0.02M & \$0.00 \\
\midrule
\textbf{Total}        &     & \textbf{5.49M} & \textbf{3.48M} & \textbf{\$41.44} \\
\bottomrule
\end{tabular}
\caption{Tool-generation cost breakdown. Token counts are from the API's reported \texttt{usage} field. GPT-5 pricing: \$1.25 / 1M input, \$10.00 / 1M output (reasoning tokens count as output). The classifier runs on a local Qwen3-VL-8B at zero API cost. Wall-clock time: 170 minutes with 4 concurrent workers.}
\label{tab:overhead}
\end{table}

At inference, the only added overhead is the per-query tool routing: a single forward pass through a sentence encoder (${\sim}$10\,ms on CPU) followed by a dot-product over the tool embedding matrix. The agent sees $|\mathcal{T}_{\text{base}}| + K = 18$ tools per query rather than the full catalog, so the per-turn context grows by only the top-$K{=}10$ generated tool schemas (${\sim}$2K tokens), a modest increase relative to the 120K context budget.

\renewcommand{\topfraction}{0.99}
\renewcommand{\textfraction}{0.01}
\setcounter{topnumber}{3}

\section{Qualitative Analysis}
\label{sec:appendix_qualitative}

We present additional case studies; three patterns account for the gains:

\begin{enumerate}[nosep]
\item \textbf{Perceive-then-query.} The generated tool chains visual identification (Lens) with structured lookup (Wikidata, text search), forcing the agent to identify the entity before searching for facts. The base agent often skips identification and searches for the wrong entity.
\item \textbf{Specialized lookup.} The generated tool directly answers a specific attribute query (e.g.\ aircraft model year, building architect) via a targeted chain-call pipeline, replacing the base agent's generic multi-step search.
\item \textbf{Forced verification.} The generated tool prevents the agent from answering from internal knowledge alone (zero-tool hallucination) by requiring a lookup step.
\end{enumerate}

Lost samples concentrate on a few tools returning incorrect information, and on mis-identification where Lens returns a similar wrong entity.

\begin{figure}[h]
\centering
\includegraphics[width=\columnwidth]{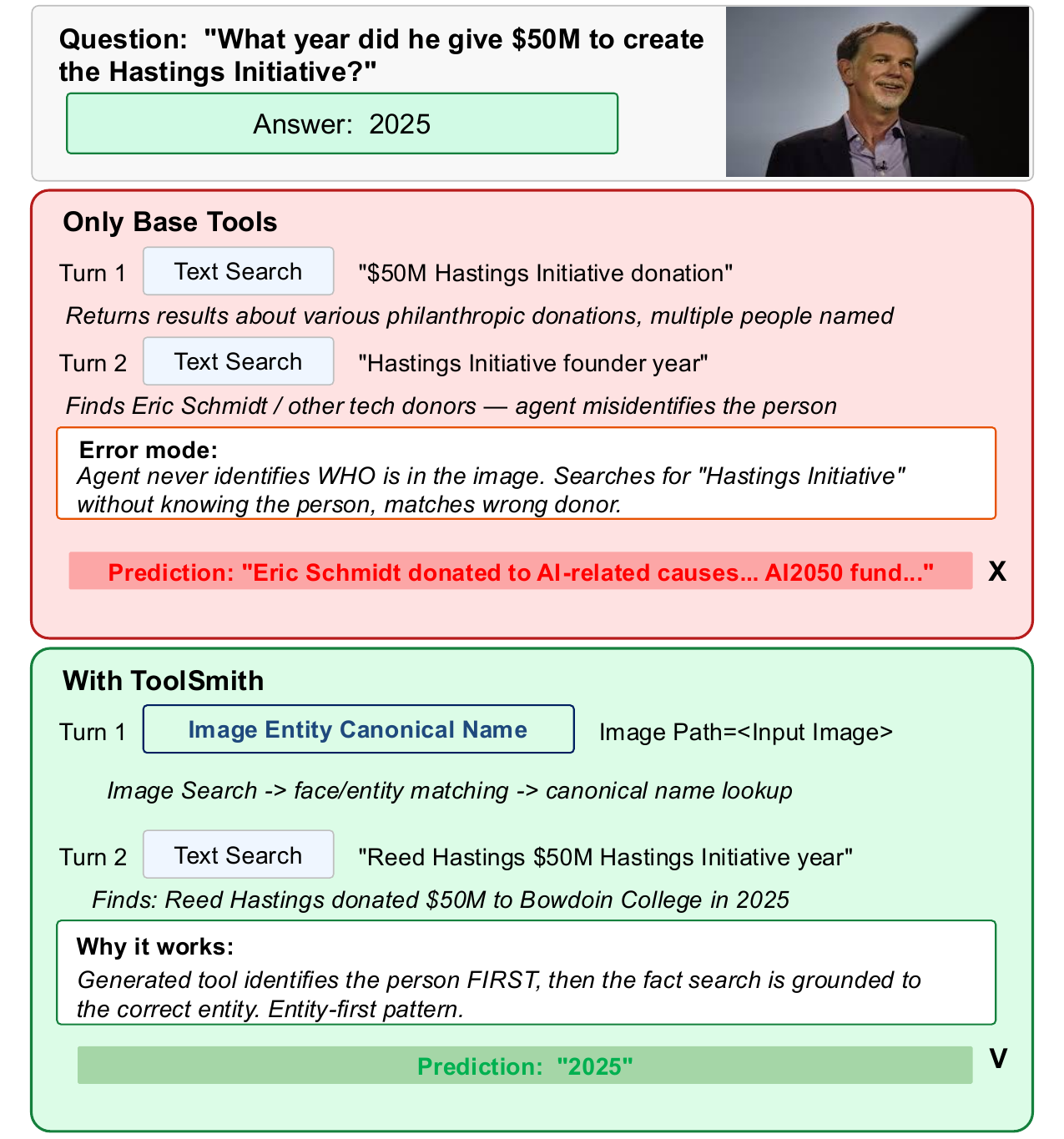}
\caption{Entity-first identification (Reed Hastings, \$50M Hastings Initiative). The base agent searches without knowing who is in the image and misattributes the donation to Eric Schmidt; the generated \toolhl{Image Entity Canonical Name} tool resolves the person first, after which a single text search returns the correct year.}
\label{fig:case_v5}
\end{figure}

\begin{figure}[h]
\centering
\includegraphics[width=\columnwidth]{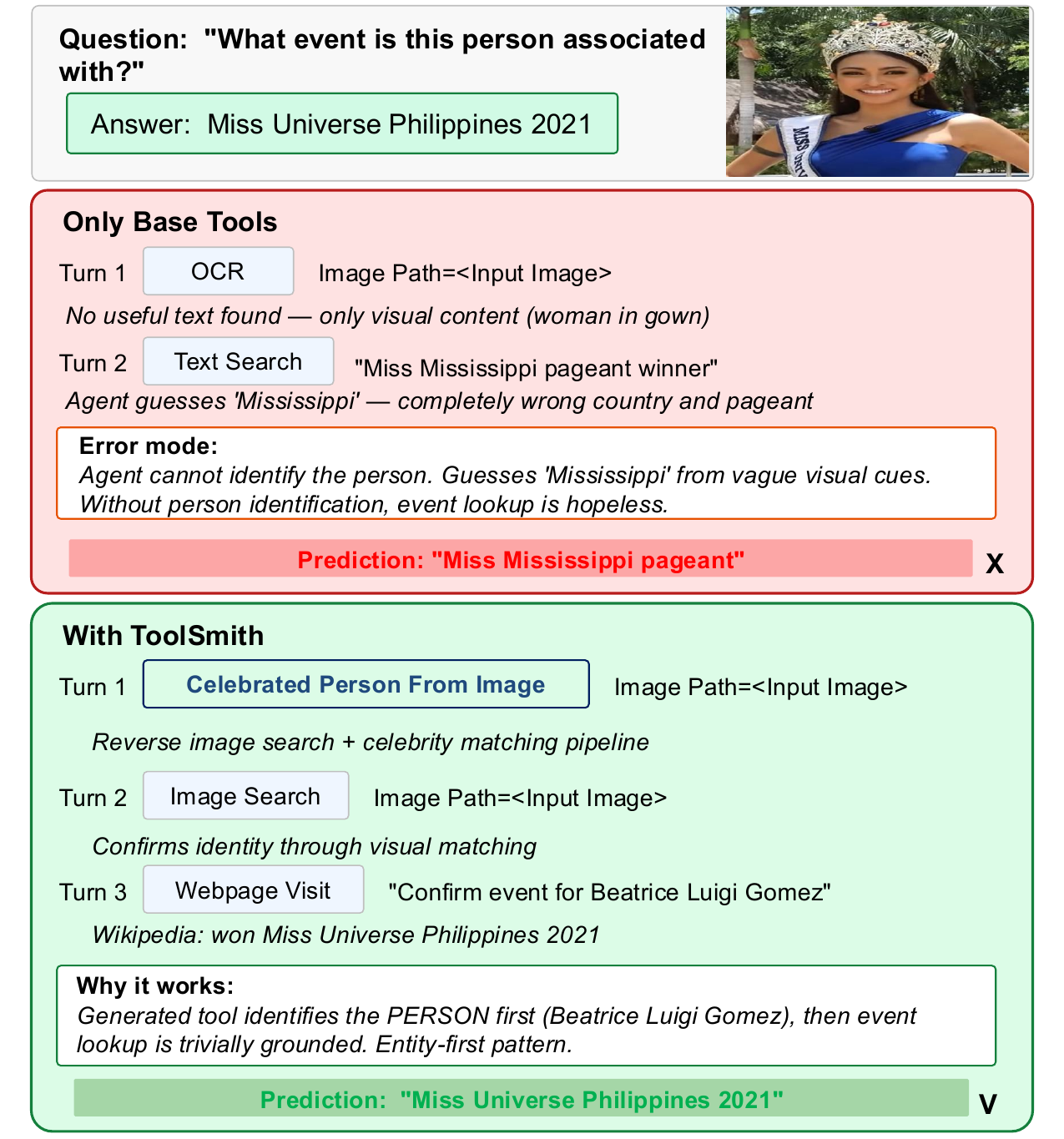}
\caption{Entity-first identification on a person whose visual context provides only vague cues. The base agent guesses ``Mississippi'' from incidental visual content and returns the wrong pageant; the generated \toolhl{Celebrated Person From Image} tool produces the canonical identity (Beatrice Luigi Gomez), which makes the event lookup trivial.}
\label{fig:case_v6}
\end{figure}

\begin{figure}[h]
\centering
\includegraphics[width=\columnwidth]{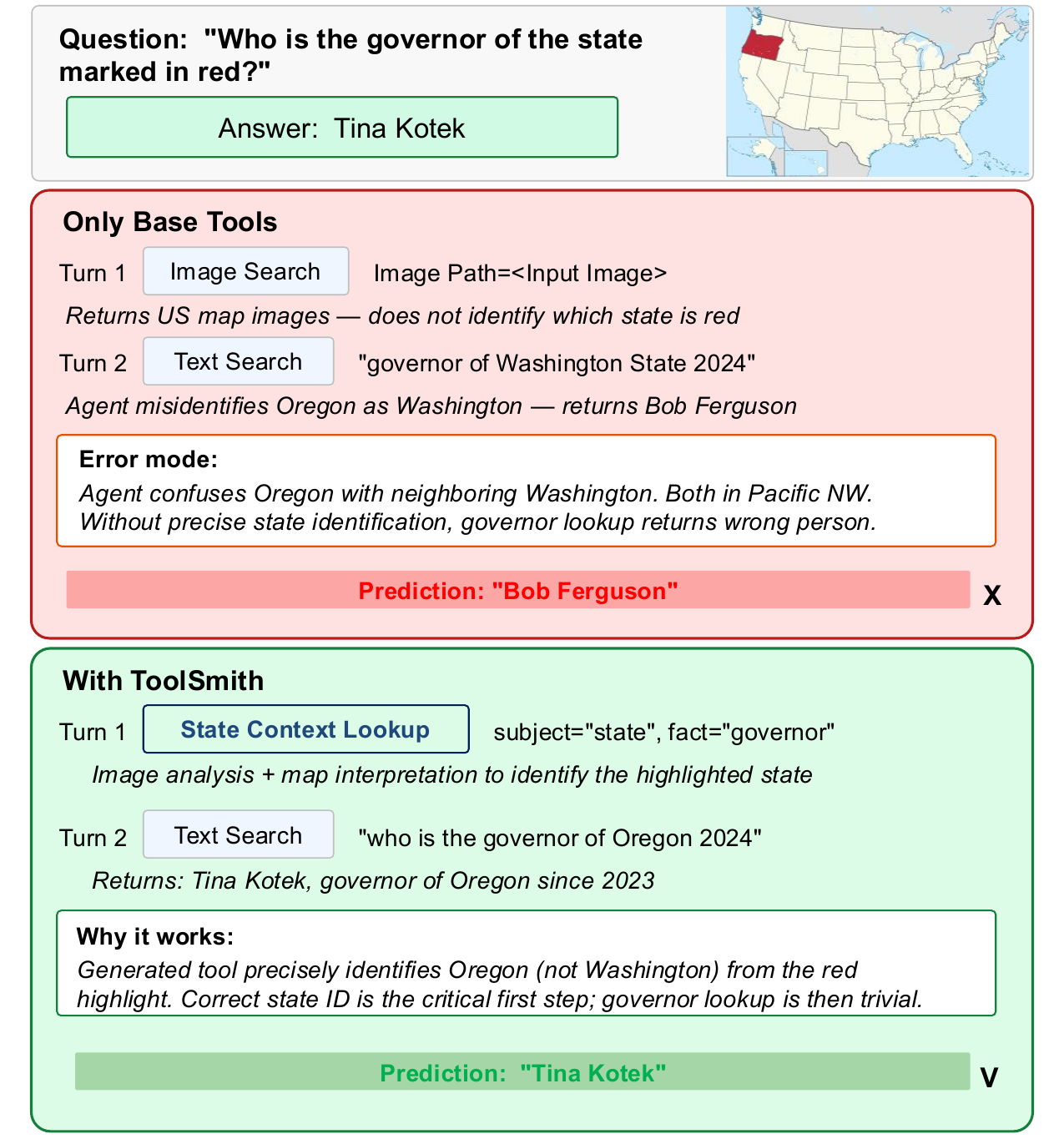}
\caption{Precise region identification on a US map. Without explicit map-region parsing the base agent confuses Oregon with neighboring Washington and returns the wrong governor; the generated \toolhl{State Context Lookup} tool pins down the highlighted state first, after which the governor lookup succeeds.}
\label{fig:case_v7}
\end{figure}

\begin{figure}[h]
\centering
\includegraphics[width=\columnwidth]{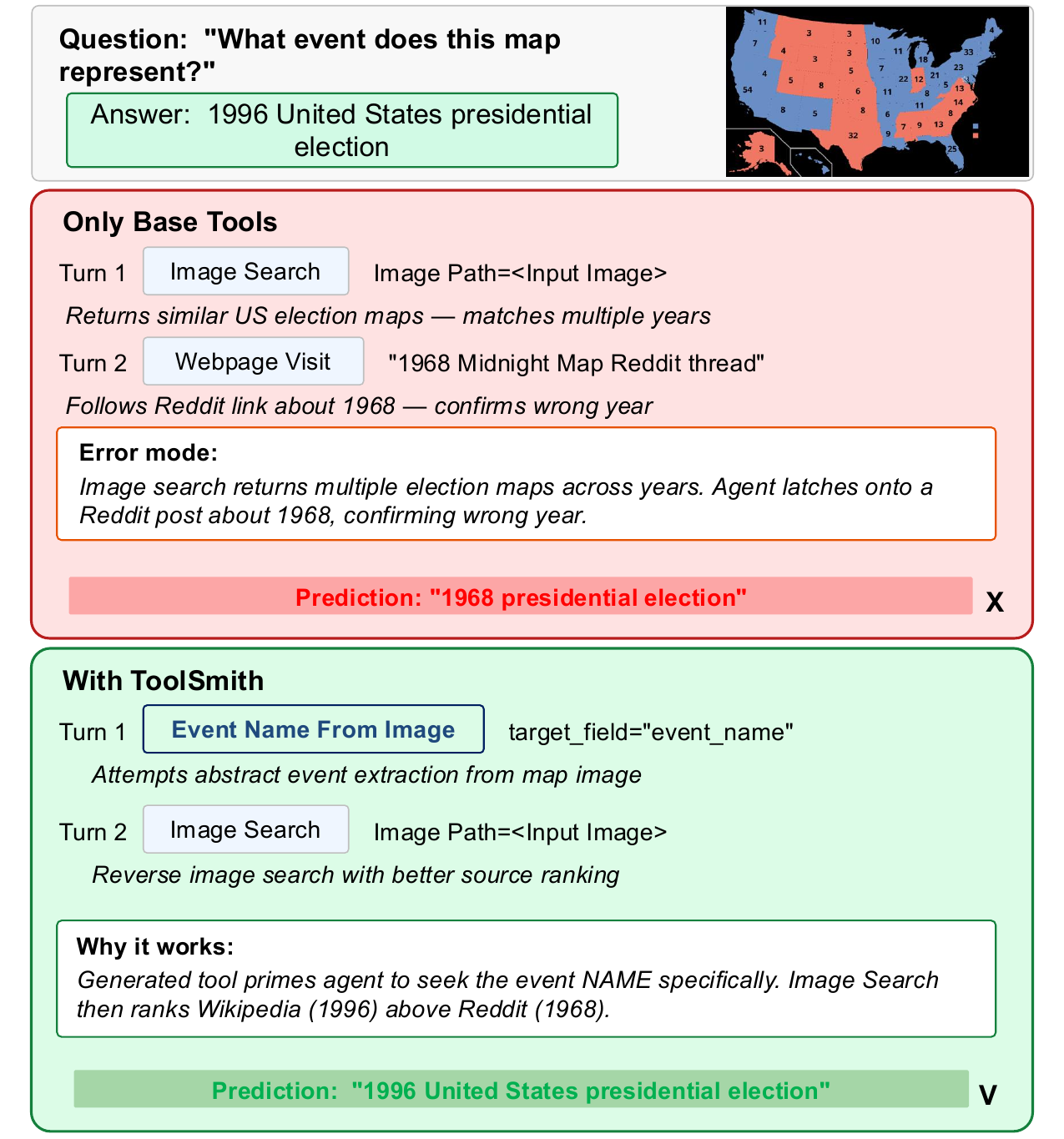}
\caption{Search-ranking steering. The base agent's generic image search returns multiple election maps and the agent latches onto a Reddit thread about the wrong year; the generated \toolhl{Event Name From Image} tool primes the search for an event-name extraction and reorders the results so Wikipedia (1996) outranks Reddit (1968).}
\label{fig:case_v8}
\end{figure}

\paragraph{Case studies.} Figures~\ref{fig:case_v5}--\ref{fig:case_v8} present four representative gained samples. The first three illustrate the perceive-then-query pattern at the entity (Figures~\ref{fig:case_v5} and~\ref{fig:case_v6}) and region (Figure~\ref{fig:case_v7}) levels; the fourth (Figure~\ref{fig:case_v8}) shows a generated tool re-ranking search results toward higher-authority sources even when both pipelines call similar primitives.

\end{document}